\documentclass[10pt,twocolumn,letterpaper]{article}

\usepackage{cvpr}
\usepackage{bbding}
\usepackage{pifont}

\newcommand{\corr}{\textsuperscript{\dag}}

\usepackage{booktabs}
\usepackage{tabularx}
\usepackage{bibunits}
\definecolor{cvprblue}{rgb}{0.21,0.49,0.74}
\usepackage[pagebackref,breaklinks,colorlinks,allcolors=cvprblue]{hyperref}

\def\paperID{*****} 
\def\confName{CVPR}
\def\confYear{2026}

\title{Learning Where to Trust: Reliability-Guided Expert Routing and Velocity Gating for 3D Deformable Registration}

\author{
Chenyu Guo$^{1,2}$,
Wenjie Dai$^1$,
Yiqian Chang$^{1,2}$
Zhiyuan Zhang$^1$,
Yunfan Wei$^1$\\
Shuhan Qi$^1$\corr,
Tianrui Xu$^3$,
Dong Liu$^3$\corr,
Ming Cong$^3$\\
$^1$ Harbin Institute of Technology, Shenzhen, Guangdong, China\\
$^2$ Peng Cheng Laboratory, Shenzhen, Guangdong, China\\
$^3$ Dalian University of Technology, Dalian, Liaoning, China\\
}

\begin{document}
\maketitle
\begingroup
\renewcommand{\thefootnote}{\dag}
\footnotetext{%
Corresponding author: \texttt{shuhanqi@cs.hitsz.edu.cn}, \texttt{liud@dlut.edu.cn}\\
Emails: \texttt{dutgcy@gmail.com}
}
\endgroup
\begin{abstract}
Deformable image registration aims to find dense anatomical correspondences between medical images. However, in low-texture, repetitive, or cross-modal regions, local feature matching often becomes ambiguous and produces multiple competing displacement candidates. Most learning-based methods ignore this spatial variation in matching reliability: they aggregate spatial context uniformly and apply full deformation updates across all voxels, which easily introduces false deformations in uncertain regions. To address this issue, we propose ReliMorph, a reliability-routed symmetric framework for 3D deformable registration that couples correspondence assessment with spatial context selection and update strength. Local correlation distributions characterize correspondence ambiguity using shared anatomical features with modality-specific adaptation. These reliability cues guide spatial expert fusion and gate residual velocities during coarse-to-fine refinement. Updated inverse-inconsistency feedback guides routing and gating at finer pyramid levels. To our knowledge, ReliMorph is the first 3D deformable registration framework to jointly condition spatial expert routing and residual velocity gating on correspondence reliability. Extensive experiments demonstrate that ReliMorph consistently outperforms competing methods in anatomical overlap, highlighting its effectiveness across unimodal and multimodal registration tasks.
\end{abstract}    
\section{Introduction}
\label{sec:intro}
Deformable image registration (DIR) estimates dense voxel-wise correspondences between medical images and plays a central role in longitudinal studies, atlas construction, image-guided intervention, and multi-modal fusion\cite{TransMorph,Transmatch,LapIRN,PIViT}. These tasks require accurate anatomical alignment while keeping the deformation field smooth and physically plausible\cite{Survey}. Classical methods solve this problem through slow iterative optimization for each image pair\cite{SyN,beg2005computing}, whereas deep learning models directly predict deformation fields in a single forward pass\cite{VoxelMorph,DALCA2019226,SYMNet}.
\begin{figure}
  \centering
  \includegraphics[width=0.47\textwidth]{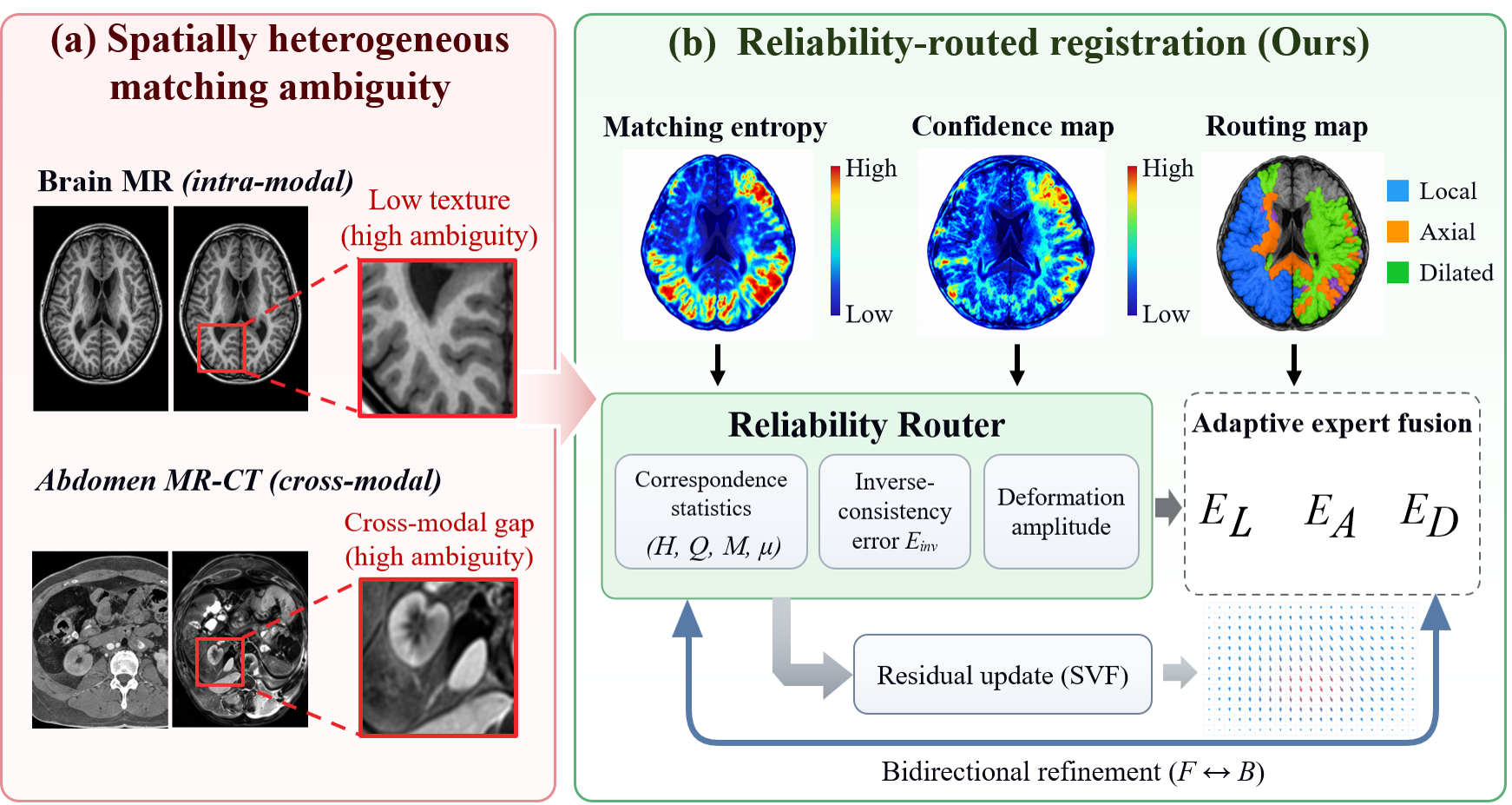}
  \caption{Motivation and validation of reliability-routed registration. (a) Conventional uniform processing applies identical context modeling and deformation updates despite spatially varying matching ambiguity. (b) In contrast, ReliMorph leverages correspondence statistics and inverse-consistency cues to adaptively route complementary experts and regulate residual updates according to matching reliability.}
  \label{Fig: Motivation}
\end{figure}

Despite fast inference, existing learning-based DIR methods still struggle when local matching becomes unreliable, facing two practical challenges:
\textbf{(1) Measuring local matching reliability.}
In homogeneous tissues, low-contrast regions, or cross-modal scans, local feature matching is often ambiguous because multiple candidate displacements yield similar correlation scores~\cite{PDD-Net,CorrMLP,li2025efficient}. Predicting a single displacement vector hides this ambiguity~\cite{DALCA2019226,9707086,rivetti2024uncertainty}. If a model trusts all local matches equally, ambiguous regions will produce noisy deformation updates; conversely, if it suppresses uncertain matches too aggressively, it may fail to register valid large deformations. We therefore need explicit cues that tell the network how reliable the local matching is at each voxel.
\textbf{(2) Adapting spatial context and update step size to reliability.}
Different regions require different receptive fields~\cite{CorrMLP,H-vit,Deformer}: sharp anatomical boundaries benefit from fine local features, whereas textureless or cross-modal regions need broader spatial context to resolve ambiguity~\cite{10656271,hu2025dual}. Recently, mixture-of-experts (MoE) models~\cite{MoE-Morph,SHMoAReg} have been used to combine multi-scale branches, and uncertainty estimation has been used to reweight training losses~\cite{Zhang2024Uncertainty}. However, loss weighting only acts during training, and feature-driven MoE routers select experts based on image appearance rather than matching quality. Even when a broader context is selected, highly ambiguous matches can still yield inaccurate velocities. Thus, a registration network should jointly decide \textit{which spatial context to use} and \textit{how much to trust the resulting velocity update} based on local matching reliability.

We propose ReliMorph, a symmetric coarse-to-fine framework for reliability-guided 3D deformable registration (Fig.~\ref{Fig: Motivation}(b)). At each pyramid level, the model assesses local matching evidence between fixed features and the currently warped moving features. This assessment guides the spatial context used to estimate residual motion and the strength of the resulting update. Broader context helps resolve competing correspondence candidates, while conservative updates limit the influence of unreliable estimates. The residual transformations are then composed with the accumulated transformations in both directions, progressively refining the alignment across scales.

Our main contributions are summarized as follows:
\begin{itemize}
    \item \textbf{Explicit correspondence reliability descriptors.} We summarize local correlation distributions using entropy, peak probability, top-2 margin, and expected offset, capturing multi-modal matching ambiguity beyond a single displacement estimate.
    \item \textbf{Reliability-guided expert routing and velocity gating.} We fuse complementary local, axial, and dilated experts using voxel-wise weights conditioned on correspondence and registration-state cues. A parallel velocity gate scales residual updates using the same conditioning input.
    \item \textbf{Cross-scale consistency feedback.} We compute forward--backward transformation disagreement at each pyramid level and feed it to the next finer level, helping the router and velocity gate detect and correct errors accumulated from coarser scales.
\end{itemize}

\begin{figure*}[t]
    \centering
    \includegraphics[width=\textwidth]{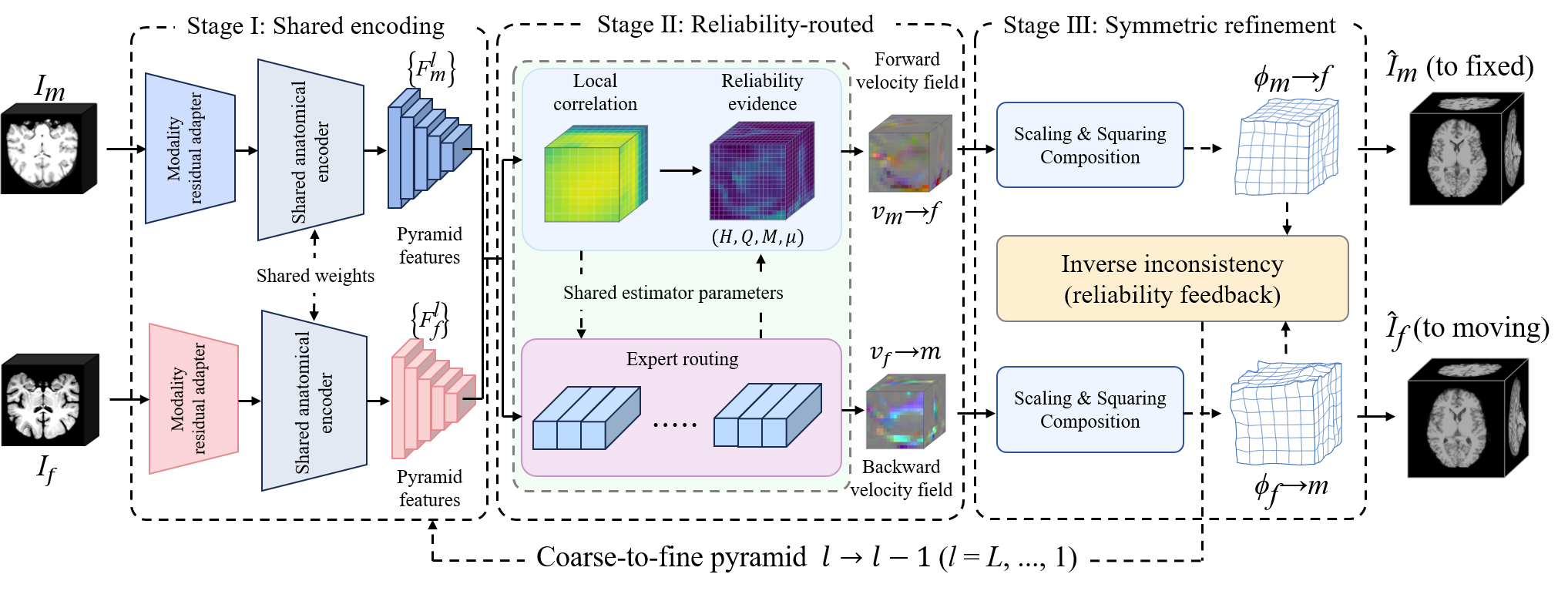}
    \caption{\textbf{Overview of ReliMorph.} \textbf{Stage I} (shared encoding) extracts multiscale anatomical features using a shared encoder with modality-specific adaptation. \textbf{Stage II} (reliability-routed estimation) uses correspondence and registration-state cues to guide expert fusion and residual-velocity gating at each pyramid level. \textbf{Stage III} (symmetric refinement) integrates gated velocities via scaling and squaring, then composes the resulting residual transformations with the accumulated transformations in both directions. Updated inverse-inconsistency maps feed back to Stage II at the next finer level.}
    \label{Fig: model}
\end{figure*}
\section{Related Work}
\label{Sec: 2}

\subsection{Cross-Modal Representations for Deformable Registration}
\label{Sec: 2.1}
Cross-modal deformable registration requires anatomical correspondence estimation despite modality-dependent differences in image appearance. Learning-based approaches address this challenge through three complementary strategies: 1) Image translation converts cross-modal alignment into within-modality matching by translating images between modalities\cite{9157219,Han2022MRCT}. 2) Appearance randomization uses synthetic training images or contrast augmentation to reduce dependence on the imaging contrasts encountered during training\cite{SynthMorph,AC-CAR}. 3) Representation learning constructs feature spaces that emphasize shared structure through contrastive learning, anatomical self-similarity, or knowledge transfer from pretrained visual encoders\cite{CoMIR,ContraReg,10656271,DINO-Reg}.

Robustness to appearance variation does not eliminate local correspondence ambiguity. Homogeneous tissues, repeated structures, and modality-dependent boundary visibility can still produce competing matches, particularly when structural overlap is incomplete.

\subsection{Coarse-to-Fine and Diffeomorphic Registration}
\label{Sec: 2.2}
Coarse-to-fine estimation and geometric regularization address complementary aspects of deformable registration. Pyramid and cascaded approaches progressively refine alignment to accommodate large deformations\cite{LapIRN,CorrMLP,RCN}. Velocity-based models construct transformations by integrating smooth velocity fields\cite{SYMNet,TransMorph}, while inverse-consistency regularization encourages agreement between forward and backward mappings\cite{ICON,GradICON}. These mechanisms can be combined to support progressive alignment and transformation regularity.

Geometric regularity and inverse consistency do not by themselves establish anatomical correctness. In progressive registration, errors in earlier updates can alter the correspondence evidence available to later stages. A remaining challenge is to use local matching reliability and bidirectional consistency feedback to guide deformation updates across scales.

\subsection{Reliability-Aware Matching and Adaptive Routing}
\label{Sec: 2.3}
Reliability assessment in matching and registration draws on probabilistic correspondence models\cite{PDC-Net+}, spatially varying variance estimates\cite{AC-CAR,Zhang2024Uncertainty}, and bidirectional transformation agreement\cite{vanharten2024robust}. These signals characterize different aspects of uncertainty or consistency, and uncertainty estimates have been used to guide regional loss weighting\cite{Zhang2024Uncertainty}. Spatially adaptive methods address how image information is processed by adjusting convolution kernels, receptive fields, or feature aggregation according to image content \cite{SACB-Net,DySNet}. Within this broader family, expert routing combines heterogeneous processing branches through input-dependent selection or weighting\cite{SHMoAReg,MoE-Morph}.

However, uncertainty-based loss weighting modifies the training objective, while content-dependent kernels and expert routing adapt feature processing to image appearance. These mechanisms alone do not specify how correspondence reliability should jointly control spatial context selection and deformation update strength.


\section{Methodology}

\subsection{Overview of ReliMorph}
\label{Sec: 3.1}
Given a moving image $I_m$ and a fixed image $I_f$, ReliMorph predicts displacement fields $u_{m\rightarrow f}$ and $u_{f\rightarrow m}$. The former is defined on the fixed-image grid, and the latter on the moving-image grid. For a source tensor $F$ and displacement field $u$, backward sampling is defined as
\begin{equation}
    \mathcal W(F,u)(p)=F(p+u(p)),
\end{equation}
where $p$ is a voxel coordinate on the output grid and $F$ is sampled using trilinear interpolation with border padding. Accordingly, $\mathcal W(I_m,u_{m\rightarrow f})$ is defined on the fixed-image grid, and $\mathcal W(I_f,u_{f\rightarrow m})$ on the moving-image grid. Displacements are expressed in voxel units of the current grid. As shown in Fig.~\ref{Fig: model}, ReliMorph consists of three stages across a coarse-to-fine feature pyramid: (1) a shared encoder with lightweight modality adapters, (2) reliability-guided expert routing and velocity gating at each scale, and (3) symmetric velocity integration with cross-scale consistency feedback.

\subsection{Shared-Anatomy Modality-Residual Encoding}
\label{Sec: 3.2}

In Stage I (Fig.~\ref{Fig: model}), the encoder uses a shared backbone to process both images while retaining lightweight adaptation to modality-dependent appearance. For each image, we construct a local intensity residual by subtracting its local mean, computed with a $5\times5\times5$ average-pooling kernel. Concatenating this residual with the original image supplies local intensity variations alongside the original appearance information.

Let $\mathbf{A}_q$ denote the concatenated input for modality $q$, and let $\mathcal{B}_0$ denote the shared shallow stem. The shallow features are computed as:
\begin{equation}
\mathbf{F}^{0}_{q}
=
\mathcal{B}_0(\mathbf{A}_q)
+
\mathcal{A}_q\!\left(\mathcal{B}_0(\mathbf{A}_q)\right),
\label{eq:modality_residual}
\end{equation}
where $\mathcal{A}_q$ is a modality-specific residual adapter. Each adapter applies depthwise spatial filtering, group normalization, a LeakyReLU activation, and a pointwise projection. The final pointwise projection is zero-initialized.

The adapted features pass through a pyramid of shared strided convolutional stages with residual blocks. Each stage is followed by modality-conditioned channel-wise affine modulation, following the feature-wise linear modulation principle~\cite{Film,AC-CAR}. Learned modality embeddings provide a residual scale and a shift, both initialized to zero. This initialization preserves the unmodulated shared computation while allowing modality-dependent adjustments to develop during training. The resulting features retain shared spatial processing throughout the hierarchy without requiring independent modality-specific encoders.

\subsection{Reliability-Routed Symmetric Correspondence Modeling}
\label{Sec: 3.3}
\begin{figure}[t]
    \centering
    \includegraphics[width=0.47\textwidth]{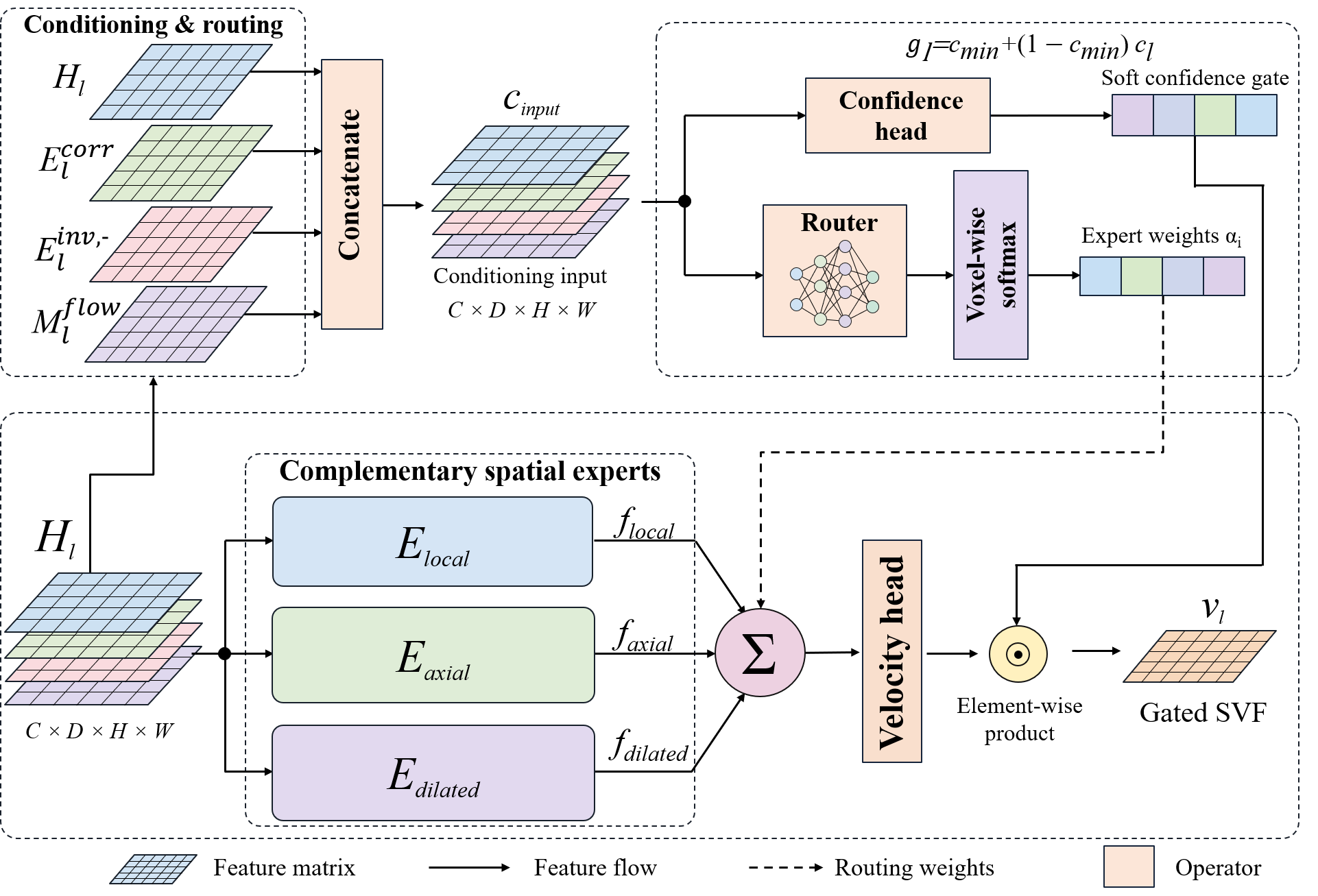}
    \caption{\textbf{Reliability-conditioned expert routing and velocity modulation.} Correspondence and registration-state cues guide voxel-wise fusion of local, axial, and dilated convolutional experts. A confidence-based gate with a positive lower bound scales the residual stationary velocity field decoded from the fused features.}
    \label{Fig: expert}
\end{figure}

In Stage II (Fig.~\ref{Fig: model}), at each pyramid level, correspondence evidence guides both spatial context aggregation and the strength of the deformation update. Figure~\ref{Fig: expert} details this module, which shares estimator parameters between the forward and backward directions.

\paragraph{Distributional local correspondence.}
We index pyramid levels from the finest level $1$ to the coarsest level $L$. At level $l$, the incoming forward displacement warps the moving features into the fixed-feature grid, yielding $\widetilde{\mathbf{F}}_m^l$. Both feature tensors are then normalized along the channel dimension. For each voxel $\mathbf{p}$, their dot products over integer offsets in $\Omega_l=[-r_l,r_l]^3\cap\mathbb{Z}^3$ form a local correlation volume $\mathbf{C}_l$\cite{CorrMLP,PDC-Net+}. Specifically, $C_l(\mathbf{p},\boldsymbol{\delta})$ compares the fixed feature at $\mathbf{p}$ with the warped moving feature at $\mathbf{p}+\boldsymbol{\delta}$. Shifted features outside the grid are zero-padded during correlation construction.

We convert these responses into a distribution over candidate offsets:
\begin{equation}
P_l(\mathbf{p},\boldsymbol{\delta})
=
\frac{\exp\!\left(C_l(\mathbf{p},\boldsymbol{\delta})/\tau_l\right)}
{\displaystyle\sum_{\boldsymbol{\delta}'\in\Omega_l}
\exp\!\left(C_l(\mathbf{p},\boldsymbol{\delta}')/\tau_l\right)}.
\label{eq:corr_prob}
\end{equation}

The level-specific temperature $\tau_l$ is learned jointly with the registration network and controls the concentration of the distribution over candidate offsets. The resulting
statistics are used as level-specific conditioning cues
for deformation estimation.

\paragraph{Correspondence reliability evidence.}
We summarize matching ambiguity using normalized entropy, peak probability, and the margin between the two largest probabilities. The normalized entropy is:
\begin{equation}
H_l(\mathbf{p})
=
-\frac{1}{\log|\Omega_l|}
\sum_{\boldsymbol{\delta}\in\Omega_l}
P_l(\mathbf{p},\boldsymbol{\delta})
\log P_l(\mathbf{p},\boldsymbol{\delta}),
\label{eq:corr_entropy}
\end{equation}
The peak probability $Q_l$ measures the mass assigned to the strongest candidate, while the top-response margin $M_l$ measures its separation from the second strongest candidate. Together with $H^l$, these statistics characterize ambiguity within the local candidate set and serve as reliability cues.

We also retain the directional tendency of the distribution through its normalized expected offset:
\begin{equation}
\boldsymbol{\mu}_l(\mathbf{p})
=
\frac{1}{r_l}
\sum_{\boldsymbol{\delta}\in\Omega_l}
P_l(\mathbf{p},\boldsymbol{\delta})\,\boldsymbol{\delta}.
\label{eq:expected_offset}
\end{equation}

Concatenating $H_l$, $Q_l$, $M_l$, and the three components of $\boldsymbol{\mu}_l$ gives the six-channel correspondence descriptor $\mathbf{E}_l^{\mathrm{corr}}$. 

\paragraph{Registration-state reliability.}
The matcher also receives the current displacement magnitude and consistency feedback from the preceding coarser level. The magnitude map $M_l^{\mathrm{flow}}$ is the voxel-wise Euclidean norm of the incoming displacement, divided by the largest spatial dimension $s_l$.

To define the consistency feedback, let superscripts $-$ and $+$ denote states before and after the update at a given level. After updating both directions, we compute the forward inverse-inconsistency map following inverse-consistent formulations\cite{ICON} as:
\begin{equation}
E_l^{\mathrm{inv},+}
=
\operatorname{clip}_{[0,1]}
\!\left(
\frac{
\left\|
\mathbf{u}^{l,+}_{m\rightarrow f}
+
\mathcal{W}\!\left(
\mathbf{u}^{l,+}_{f\rightarrow m},
\mathbf{u}^{l,+}_{m\rightarrow f}
\right)
\right\|_2
}{s_l}
\right).
\label{Eq: inverse_error}
\end{equation}

The forward inverse-consistency residual is evaluated on the fixed-image grid. The reverse residual is evaluated on the moving-image grid by exchanging the two displacement fields. At the coarsest level, the incoming feedback $E_L^{\mathrm{inv},-}$ is initialized to zero. At finer levels, $E_l^{\mathrm{inv},-}$ is obtained by trilinearly interpolating the updated map from the preceding coarser level. This normalized scalar feedback is interpolated without additional magnitude rescaling.

We concatenate the fixed features, warped moving features, their absolute difference, the correlation volume, and the correspondence and registration-state descriptors. A convolutional projection followed by a residual block produces the matching representation \(H_l\).

The router and confidence head additionally receive the correspondence and registration-state descriptors through direct concatenation with \(H_l\).
\begin{equation}
    \boldsymbol{\alpha}_l = \mathrm{Softmax}_k\Big(\mathcal{R}_l\big(\mathrm{Cat}[\mathbf{H}_l, \mathbf{E}_l^{\mathrm{corr}}, E_l^{\mathrm{inv},-}, M_l^{\mathrm{flow}}]\big)\Big).
    \label{eq:routing}
\end{equation}

The router predicts voxel-wise fusion weights for three experts with distinct spatial sampling patterns. The local-detail expert uses a \(3\times3\times3\) kernel to aggregate nearby features. The axial-context expert sums responses from \(7\times1\times1\), \(1\times7\times1\), and \(1\times1\times7\) kernels, extending context along the three coordinate axes. The dilated-context expert uses a \(3\times3\times3\) kernel with dilation rate 2 to sample a broader three-dimensional neighborhood. Each expert applies depthwise spatial filtering followed by a \(1\times1\times1\) projection. The learned routing weights adapt their relative contributions at each voxel according to correspondence and registration-state evidence. We fuse the three expert outputs using these voxel-wise weights:
\begin{equation}
\widetilde{\mathbf{H}}_l
=
\sum_{k\in\{\mathrm{loc},\mathrm{axial},\mathrm{dil}\}}
\alpha_l^k\odot\mathcal{E}_{l,k}(\mathbf{H}_l),
\label{eq:routed_experts}
\end{equation}
where $\odot$ denotes element-wise multiplication with broadcasting over feature channels. 

\noindent\textbf{Reliability-conditioned velocity modulation.}
\noindent\textbf{Reliability-conditioned velocity modulation.}
A velocity head $V_l$ decodes the routed features into a residual stationary velocity field. In parallel, a sigmoid confidence head produces a modulation score $c_l\in[0,1]$ from the same conditioning input as the router. We map this score to the effective velocity gate $g_l=c_{\min}+(1-c_{\min})c_l$, yielding:
\begin{equation}
\mathbf{v}_l
=
g_l\odot\mathcal{V}_l(\widetilde{\mathbf{H}}_l).
\label{Eq: velocity_modulation}
\end{equation}

With $c_{\min}=0.25$, gating alone retains at least one
quarter of the raw velocity magnitude without enforcing
nonzero motion.

\subsection{Symmetric Coarse-to-Fine Diffeomorphic Refinement}
\label{Sec: 3.4}

\paragraph{Residual velocity integration and composition.}
In Stage III (Fig.~\ref{Fig: model}), we construct each residual transformation by integrating its stationary velocity field. Following the scaling-and-squaring formulation used in diffeomorphic registration~\cite{DALCA2019226}, the residual displacement is
\begin{equation}
\Delta\mathbf{u}_{m\rightarrow f}^{l}
=
\exp\!\left(\mathbf{v}_{m\rightarrow f}^{l}\right)
-
\operatorname{Id},
\label{eq:velocity_exp}
\end{equation}
where $\exp$ denotes the unit-time flow of the stationary velocity and $\operatorname{Id}$ is the identity transformation. Numerically, we divide the velocity by $2^T$ and perform $T$ self-compositions using differentiable spatial warping, with $T=7$ by default. The same procedure integrates the backward velocity into its residual displacement.

We then compose the residual with the accumulated displacement:
\begin{equation}
\mathbf{u}_{m\rightarrow f}^{l,+}
=
\Delta\mathbf{u}_{m\rightarrow f}^{l}
+
\mathcal{W}\!\left(
\mathbf{u}_{m\rightarrow f}^{l,-},
\Delta\mathbf{u}_{m\rightarrow f}^{l}
\right).
\label{eq:flow_composition}
\end{equation}

This ordering evaluates the accumulated transformation after the residual transformation, accounting for the coordinates changed by the current update. The backward displacement is updated through the same composition rule.

\paragraph{Coarse-to-fine propagation and consistency feedback.}
Both displacement fields start from zero at the coarsest pyramid level. After each update, trilinear interpolation transfers the fields to the next finer grid. Each displacement component is also rescaled by the ratio of the new to old grid interval counts along its corresponding axis. The resized fields warp the next pair of features before the matcher predicts further residual velocities.

After both directions are updated, the inverse-inconsistency
maps in Eq.~\eqref{Eq: inverse_error} are interpolated to the next finer level to condition its router and confidence head.

After the finest-level update, both displacement fields are resized to the original image resolution using the same component-wise rescaling rule. The moving and fixed images are then warped with their corresponding displacement fields to produce the registered outputs. For each direction $d$, the model returns a final confidence map by combining the finest-level modulation score with the post-update inverse-inconsistency map:
\[
\hat c_d
=
U\!\left[
c_{1,d}\left(1-E_{1,d}^{\mathrm{inv},+}\right)
\right],
\]
where $U$ denotes trilinear upsampling to the original image grid.

Here, symmetry refers to shared estimation parameters and bidirectional feedback. Exact inverse consistency and fold-free discrete transformations are not guaranteed.

\begin{table*}[t]
\centering
\caption{Registration performance on IXI, OASIS, and AbdomMRCT. HD95 is measured in millimeters. Fold denotes the percentage of voxels with $\det(\nabla\phi) \leq 0$, reported to assess transformation regularity.}
\label{Tab: reg}

\small
\setlength{\tabcolsep}{0pt}
\renewcommand{\arraystretch}{1.08}

\begin{tabular*}{\textwidth}{@{\extracolsep{\fill}}l*{10}{c}@{}}
\toprule

& \multicolumn{3}{c}{IXI}
& \multicolumn{3}{c}{OASIS}
& \multicolumn{4}{c}{AbdomMRCT} \\

\cmidrule(lr){2-4}
\cmidrule(lr){5-7}
\cmidrule(lr){8-11}

Method
& DSC $\uparrow$
& HD95 $\downarrow$
& Fold $\downarrow$
& DSC $\uparrow$
& HD95 $\downarrow$
& Fold $\downarrow$
& DSC $\uparrow$
& HD95 $\downarrow$
& Fold $\downarrow$
& $\mathcal{L}_{\mathrm{MIND\text{-}SSC}}
   \downarrow$ \\

\midrule

SyN
& 0.7460 & 3.157 & 0
& 0.7454 & 2.350 & 0
& 0.8212 & \underline{8.746} & 0 & 0.0894 \\

TransMorph
& 0.7934 & 2.257 & 0.2879
& 0.7965 & 2.009 & 0.7456
& \underline{0.8798} & 9.542 & 20.735 & \underline{0.0315}\\

CorrMLP
& \underline{0.8070} & \underline{2.136} & 0.4508
& 0.8017 & 1.855 & 0.0115
& 0.8591 & 18.517 & 17.741 & 0.0412\\

AC-CAR
& 0.7857 & 2.363 & 0.1860
& 0.8080 & 1.895 & 0.1819
& 0.8557 & 13.124 & 18.020 & 0.0445\\

SACB-Net
& 0.7911 & 2.327 & 0.0189
& 0.8132 & 1.824 & 0.6495
& 0.8588 & 13.553 & 4.7408 & 0.0426 \\

ShiftMorph
& 0.7867 & 2.763 & 0.2967
& 0.7823 & 2.064 & 0.4163
& 0.8675 & 9.732 & 0.2596 & 0.1270 \\

LapIRN (SVF)
& 0.7522 & 2.930 & 0
& 0.7472 & 2.649 & 0
& 0.8098 & 13.88 & 0 & 0.1270 \\

SYMNet
& 0.7625 & 2.815 & 0.0003
& 0.7882 & 2.503 & 0.0003
& 0.7907 & 13.42 & 1.5997 & 0.1384 \\

LGANet++
& 0.7962 & 2.262 & 0.0001
& \underline{0.8164} & \underline{1.808} & 0.0013
& 0.8611 & 13.69 & 0.7292 & 0.0348 \\

\midrule

\textbf{Ours}
& \textbf{0.8093} & \textbf{2.040} & 0.0032
& \textbf{0.8258} & \textbf{1.709} & 0.0423
& \textbf{0.8868} & \textbf{6.970} & 0.0290 & \textbf{0.0284} \\

\bottomrule
\end{tabular*}
\end{table*}

\subsection{Training Objective}
\label{Sec: 3.5}
ReliMorph is trained without segmentation supervision using
image alignment, geometric regularization, and confidence
self-supervision:
\begin{equation}
\begin{aligned}
\mathcal{L} ={}&
\mathcal{L}_{\mathrm{sim}}
+\lambda_{\mathrm{ms}}\mathcal{L}_{\mathrm{ms}}
+\lambda_{\mathrm{str}}\mathcal{L}_{\mathrm{str}} \\
&+\lambda_{\mathrm{vel}}\mathcal{L}_{\mathrm{vel}}
+\lambda_{\mathrm{inv}}\mathcal{L}_{\mathrm{inv}}
+\lambda_{\mathrm{jac}}\mathcal{L}_{\mathrm{jac}}
+\lambda_{\mathrm{conf}}\mathcal{L}_{\mathrm{conf}}.
\end{aligned}
\end{equation}

The alignment terms supervise full-resolution and multiscale registration while encouraging structural agreement. Geometric regularization promotes smooth residual velocities, inverse consistency, and positive Jacobian determinants. The confidence term supervises the final consistency-adjusted maps $\hat c_d$ using soft targets $\exp[-\operatorname{sg}(e_d)/t_c]$ derived from local registration dissimilarity $e_d$. Here, $\operatorname{sg}$ stops gradients through the targets, and $t_c$ controls their decay. The targets assign higher scores to smaller local image discrepancies.

\begin{figure*}[t]
    \centering
    \includegraphics[width=\textwidth]{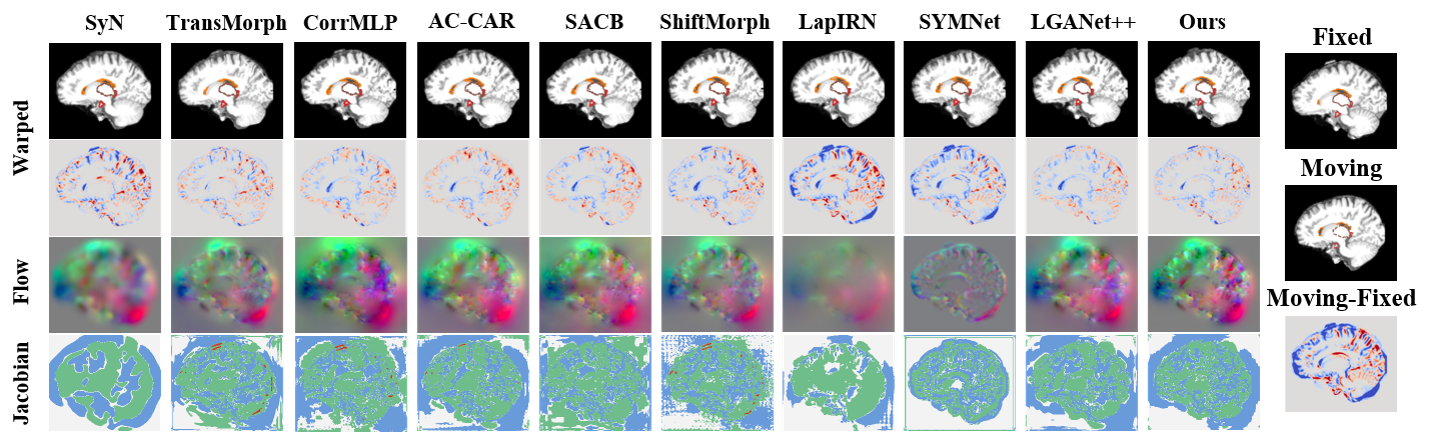}
    \caption{
      Qualitative comparison of inter-subject brain MRI registration on IXI. For each method, rows show the warped moving image with selected anatomical contours, the signed intensity residual $I_{\mathrm{warp}}-I_{\mathrm{fixed}}$, the displacement field (RGB encodes $d_x,d_y,d_z$), and the Jacobian determinant $J$. The panels on the right show the fixed and moving images and their intensity difference before registration. Residual maps share the color scale shown below. In the Jacobian maps, red indicates folding ($J\leq 0$), gray indicates near-unit volume change ($|J-1|\leq 0.02$), green indicates compression, and blue indicates expansion.
    }
    \label{Fig: IXI_results}
\end{figure*}

\section{Experiment}
\subsection{Experimental Setup}
\label{Sec: 4.1}

\noindent\textbf{Datasets.}
We evaluate ReliMorph on three public datasets covering
inter-subject brain MR and abdominal CT-MR registration.
For each brain dataset, training randomly samples 200
inter-subject pairs per epoch. All evaluations use fixed
image pairs.

\begin{enumerate}
    \item \textit{IXI}~\cite{IXI} For T1-weighted brain MR registration, we use subject-disjoint training, validation, and test sets of 403, 58, and 115 subjects, respectively. Validation and testing use 58 and 115 circular pairs. Evaluation covers 28 structures with input volumes of $160\times192\times224$.

    \item \textit{OASIS}~\cite{OASIS} We use the Learn2Reg OASIS release for inter-subject T1-weighted brain MR registration with strictly subject-disjoint training, validation, and test splits. Validation pairs are used solely for checkpoint selection, and final evaluation is reported on the held-out test pairs across 35 anatomical structures using input volumes of $192 \times 224 \times 160$.

    \item \textit{AbdomMRCT} from
    Learn2Reg~\cite{Learn2Reg} We perform intra-patient registration from moving CT to fixed T1-weighted MR using the official pairing manifests. Evaluation covers the liver, spleen, and both kidneys on $192\times160\times192$ volumes resampled to $2\,\mathrm{mm}$ isotropic spacing.
\end{enumerate}

\noindent\textbf{Comparisons.}
We select baselines covering four complementary aspects:
correspondence and context modeling (TransMorph~\cite{TransMorph}, CorrMLP~\cite{CorrMLP}, and LGANet++~\cite{LGANet++}); adaptive feature processing (AC-CAR~\cite{AC-CAR} and SACB-Net~\cite{SACB-Net}); diffeomorphic modeling (SyN~\cite{SyN}, LapIRN's SVF variant~\cite{LapIRN}, and SYMNet~\cite{SYMNet}); and efficient convolutional registration (ShiftMorph~\cite{ShiftMorph}).

Supplementary comparisons with UTSRMorph\cite{UTSRMorph} and TransMatch\cite{Transmatch} extend coverage of feature representation and explicit matching, while DySNet\cite{DySNet} and SHMoAReg\cite{SHMoAReg} provide additional references for dynamic spatial processing and expert routing. All methods share the data and evaluation protocols, with method-specific training settings, geometric regularizers, and cross-modal adaptations detailed in the supplementary material.

\noindent\textbf{Implementation and evaluation metrics.}
Our method is implemented in PyTorch and trained for
100 epochs on an NVIDIA A6000 GPU (48 GB), with a batch
size of 1. We use AdamW with an initial learning rate
and weight decay of $10^{-4}$, together with cosine
learning-rate annealing. Segmentation labels are excluded from the training loss, and checkpoints are selected by mean validation DSC, and all test sets remain strictly held out for final evaluation. Baselines use either the same hyperparameters or the configurations reported in their original papers.

We evaluate anatomical overlap using DSC and boundary
discrepancy using HD95 (mm).
Deformation regularity is assessed using Fold,
the percentage of voxels with
$\det(\nabla\phi)\leq 0$.
Anatomical metrics exclude the background and are
averaged over the evaluated structures and image pairs.
For AbdomMRCT, we additionally report
$\mathcal{L}_{\mathrm{MIND\text{-}SSC}}$,
the mean squared MIND-SSC descriptor difference
between the fixed and registered images,
to assess modality-independent structural agreement.
Lower values indicate better agreement.

\noindent\textbf{Training loss configuration.}
Full-resolution and multiscale alignment use one minus squared LNCC for brain MR and mean-squared MIND-SSC descriptor differences for abdominal CT--MR. Multiscale weights are normalized and double from coarser to finer levels. Structural alignment applies Smooth-$L_1$ loss to spatially mean-normalized gradient magnitudes, computed after $5\times5\times5$ local-mean subtraction.

Velocity regularization averages squared spatial differences
across levels and directions. Inverse-consistency regularization uses mean-squared bidirectional composition residuals in voxel coordinates, without the normalization or clipping used for routing feedback.

The Jacobian penalty averages squared negative Jacobian
determinants. Confidence supervision uses bidirectionally averaged Smooth-$L_1$ loss. Loss weights are reported in Tab.~\ref{Tab: S1}, with $\lambda_{\mathrm{str}}=0$ for AbdomMRCT.

\subsection{Performance Comparison}
\label{Sec: 4.2}
\textbf{Quantitative comparison.}
Tab.~\ref{Tab: reg} shows that ReliMorph achieves the highest DSC and lowest HD95 across all three benchmarks. On IXI, it improves DSC from 0.8070 to 0.8093 and reduces HD95 from 2.136 to 2.040 mm relative to CorrMLP. On OASIS, it improves DSC from 0.8164 to 0.8258 and reduces HD95 from 1.808 to 1.709 mm relative to LGANet++. Its folding rate is higher than that of LGANet++ (0.0423\% vs. 0.0013\%). On AbdomMRCT, ReliMorph achieves a DSC of 0.8868 and an HD95 of 6.970 mm, compared with 0.8798 and 9.542 mm for TransMorph. Its folding rate and MIND-SSC dissimilarity are also lower than those of TransMorph.

\begin{figure*}[t]
    \centering
    \includegraphics[width=\textwidth]{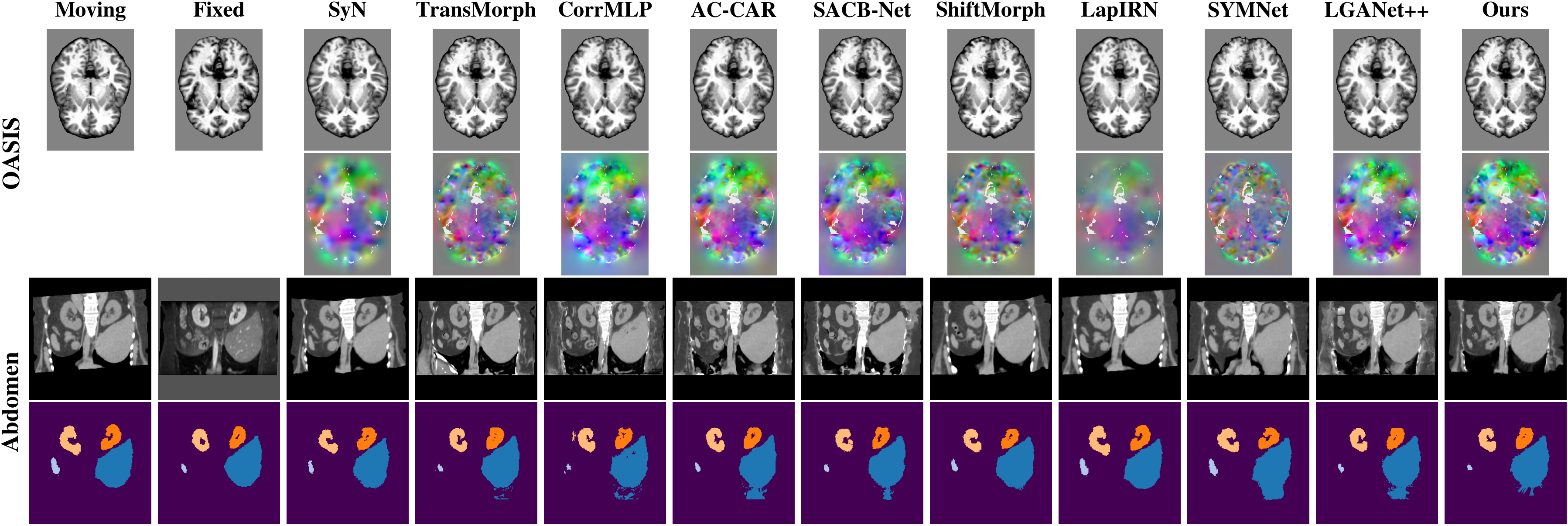}
    \caption{
    \textbf{Qualitative comparison of deformable registration on OASIS and abdominal images.} The first two columns show the moving and fixed images; the remaining columns show results from the compared methods. For OASIS, the top two rows show the warped images and color-coded deformation fields. For the abdominal case, the bottom two rows show the warped images and corresponding propagated segmentation masks.
    }
    \label{Fig: OASIS}
\end{figure*}

\textbf{Qualitative comparison.}
As shown in Fig.~\ref{Fig: IXI_results} and Fig.~\ref{Fig: OASIS}, we further provide qualitative comparisons on IXI, OASIS, and AbdomMRCT.
In the displayed IXI case, ReliMorph shows weaker intensity residuals around cortical boundaries than several baselines while maintaining smooth Jacobian maps.
The OASIS result closely follows the fixed-image anatomical layout, while the abdominal propagated labels show accurate organ-level alignment with less pronounced elongation and fragmentation than competing methods, benefiting from routing ambiguous boundaries toward broader spatial contexts.

\begin{figure}[t]
    \centering
    \includegraphics[width=0.45\textwidth]{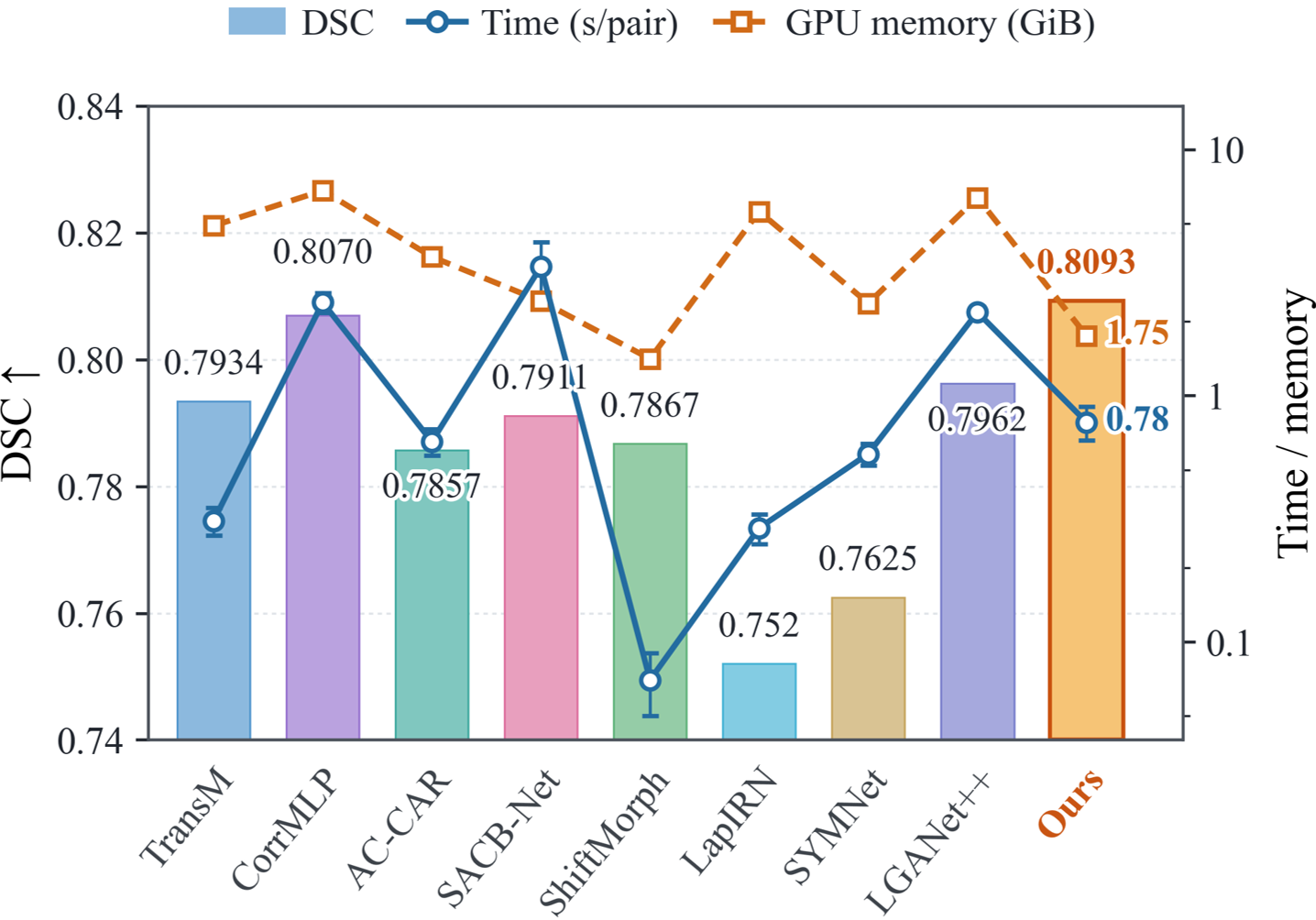}
    \caption{Comparison of registration accuracy (DSC), inference time, and peak GPU memory usage across methods on the IXI dataset.}
    \label{Fig: param}
\end{figure}

\textbf{Computational efficiency.}
Finally, Fig.~\ref{Fig: param} compares registration accuracy, inference time, and peak GPU memory on IXI.
By using compact reliability descriptors and lightweight depthwise-separable experts, ReliMorph achieves a DSC of 0.8093 with inference taking $0.78 \pm 0.12$~s per image pair and requiring only 1.75~GiB of peak allocated GPU memory.

\subsection{Ablation Study}

\noindent\textbf{Disentangling reliability-conditioned routing and gating.}
Tab.~\ref{Tab: ablation} evaluates direct descriptor conditioning of expert routing and velocity gating on IXI. With \(H_l\) held fixed, direct descriptor conditioning increases DSC from 0.7894 to 0.7986 for routing alone and to 0.7972 for gating alone. Conditioning both components directly on the descriptors achieves a DSC of 0.8093.

\begin{table}[t]
    \centering
    \caption{Ablation study of ReliMorph on IXI. Feat. denotes conditioning on \(H_l\) alone, whereas Rel. denotes conditioning on \(H_l\) concatenated with explicit correspondence and registration-state descriptors. Across the \(2\times2\) routing–gating conditioning comparison, \(H_l\) is constructed identically and includes these descriptors.}
    \label{Tab: ablation}
    \footnotesize
    \setlength{\tabcolsep}{2.8pt}
    \renewcommand{\arraystretch}{1.10}

    \begin{tabularx}{\columnwidth}{
        @{}>{\raggedright\arraybackslash}Xccc@{}
    }
        \toprule
        Variant
        & DSC $\uparrow$
        & HD95 $\downarrow$
        & Fold $\downarrow$ \\
        &  & (mm) & (\%) \\
        \midrule
        \textbf{Full model (Rel.\ routing + Rel.\ gating)} & \textbf{0.8093} & \textbf{2.040} & 0.0032 \\
        \midrule
        \multicolumn{4}{@{}l}{\textit{Routing \& gating conditioning cross-ablation}} \\
        Feature-only MoE (no $\mathbf{C}_l$ or reliability stats) & 0.7852 & 2.381 & 0.0002 \\ 
        Feat.\ routing + Feat.\ gating (raw $\mathbf{C}_l$ in $\mathbf{H}_l$) & 0.7894 & 2.336 & 0.0002 \\ 
        Rel.\ routing + Feat.\ gating & 0.7986 & 2.204 & 0.0014 \\ 
        Feat.\ routing + Rel.\ gating & 0.7972 & 2.221 & 0.0011 \\ 
        \midrule
        \multicolumn{4}{@{}l}{\textit{Reliability descriptors \& module ablation}} \\
        w/o ambiguity statistics    & 0.7911 & 2.317 & 0.0001 \\
        w/o correlation descriptors & 0.7875 & 2.356 & 0.0001 \\
        w/o consistency feedback    & 0.7874 & 2.340 & 0.0001 \\
        Uniform expert fusion       & 0.7907 & 2.312 & 0.0001 \\
        w/o velocity gating         & 0.7919 & 2.314 & 0.0001 \\
        w/o positive gate floor     & 0.7745 & 2.460 & 0 \\
        \bottomrule
    \end{tabularx}

    \par\vspace{2pt}
    \parbox{\columnwidth}{\footnotesize
        Fold denotes the percentage of voxels with
        $\det(\nabla\phi)\leq 0$.
    }
\end{table}

\noindent\textbf{Reliability assessment and expert specialization.}
For Tab.~\ref{Tab: expert_specialization}, we divide fixed-image foreground voxels into low-, medium-, and high-ambiguity groups using per-pair tertiles of finest-level matching entropy. Entropy is measured before the finest-level residual update, and group membership remains fixed across expert interventions. Anatomical label agreement is 93.8\%, 86.4\%, and 74.2\% across the three groups, respectively.

In repetitive cortical folds, the false-match rate among the top 20\% most confident predictions is 8.4\% when using peak probability $Q_l$ alone. Including the top-2 margin $M_l$ and entropy $H_l$ reduces this rate to 5.6\%, while further incorporating bidirectional inconsistency $E_l^{\mathrm{inv},-}$ reduces it to 3.1\%. Additional reliability evaluations are provided in Sec.~\ref{Sec: supp_conf_error} of the supplementary material.

\noindent\textbf{Expert complementarity and specialization.} The heterogeneous expert configuration achieves an IXI DSC of 0.8093, compared with 0.7873 for three independently parameterized local experts (Tab.~\ref{tab:supp_expert_composition}). Both configurations have a reported parameter count of 0.431M. Replacing learned routing with uniform fusion reduces DSC to 0.7907 (Tab.~\ref{Tab: ablation}). In low-, medium-, and high-ambiguity regions, the largest mean routing weights are assigned to the local (0.468), axial (0.369), and dilated (0.428) experts, respectively (Tab.~\ref{Tab: expert_specialization}). Suppressing these experts at the finest level of the frozen model increases anatomical label disagreement by 1.42, 1.09, and 1.56 percentage points in the corresponding regions.

\begin{table}[t]
\centering
\caption{Reliability calibration and expert specialization across ambiguity regimes on IXI. Acc.\ denotes actual regional anatomical correspondence accuracy (\%). $\Delta e$ denotes the increase in regional anatomical label disagreement (pp) after suppressing each expert at the finest matching level of the frozen model.}
\label{Tab: expert_specialization}
\footnotesize
\setlength{\tabcolsep}{2.6pt}
\renewcommand{\arraystretch}{1.08}
\begin{tabular}{@{}lccccccc@{}}
\toprule
& & \multicolumn{3}{c}{Mean routing weight}
& \multicolumn{3}{c}{$\Delta e$ (pp)} \\
\cmidrule(lr){3-5}
\cmidrule(lr){6-8}
Ambiguity & Acc.\ (\%) $\uparrow$ & Local & Axial & Dilated
          & Local & Axial & Dilated \\
\midrule
Low    & 93.8 & 0.468 & 0.314 & 0.218
       & +1.42 & +0.76 & +0.38 \\ 
Medium & 86.4 & 0.338 & 0.369 & 0.293
       & +0.85 & +1.09 & +0.74 \\ 
High   & 74.2 & 0.219 & 0.353 & 0.428
       & +0.41 & +1.14 & +1.56 \\ 
\bottomrule
\end{tabular}
\end{table}

\section{Conclusion}
We presented ReliMorph, a symmetric coarse-to-fine 3D registration framework that explicitly uses matching reliability to guide where to aggregate spatial context and how strongly to update deformations. By summarizing local correlation distributions with entropy, peak probability, top-2 margin, and expected offset—and combining them with cross-scale inverse-inconsistency feedback—ReliMorph routes ambiguous regions toward broader receptive fields while scaling down unreliable velocity updates. Extensive experiments on brain MR and cross-modal abdominal CT--MR benchmarks show that ReliMorph consistently outperforms state-of-the-art registration methods in both alignment accuracy and deformation regularity.

\clearpage
    {
    \small
    \bibliographystyle {ieeenat_fullname}
    \bibliography {main}
    }

\clearpage
\setcounter{page}{1}
\maketitlesupplementary

\appendix
\setcounter{section}{0}
\renewcommand{\thesection}{A\arabic{section}}
\renewcommand{\thesubsection}{\thesection.\arabic{subsection}}

\setcounter{figure}{0}
\renewcommand{\thefigure}{S\arabic{figure}}

\setcounter{table}{0}
\renewcommand{\thetable}{S\arabic{table}}

\setcounter{equation}{0}
\renewcommand{\theequation}{S\arabic{equation}}

\section{Reproducibility Details}
\label{Sec: supp_reproducibility}
For IXI and OASIS, pre-alignment and intensity normalization follow the TransMorph FreeSurfer pipeline\cite{TransMorph} (skull-stripping, affine alignment, with intensities scaled to $[0,1]$) and the Learn2Reg/HyperMorph\cite{Learn2Reg} OASIS preprocessing pipeline with foreground percentile clipping at $0.5$–$99.5\%$ followed by $z$-score normalization, respectively. For AbdomMRCT, affine initialization is obtained from a 12-DoF affine transformation optimized by coarse-to-fine Adam, where the loss consists of MIND-SSC distance within overlapping body masks plus a soft body-mask Dice term. CT intensities are processed by clipping to $[-200,400]$ HU and min–max scaling to $[0,1]$, and MR intensities use the same $0.5$–$99.5\%$ percentile clipping and $z$-score normalization.

Table~\ref{Tab: S1} lists the remaining hyperparameters. Search radii are measured in feature-grid voxels. The learnable temperature is initialized as $\tau^{(0)}=0.07$ and parameterized in the log domain as $\tau=\exp(\log\tau)$, with values clamped to $[0.02,0.50]$. Unless otherwise specified, retrained ablations use the default settings in Table~\ref{Tab: S1} and the dataset-specific training protocol described in the main paper. Within each dataset, variants are trained independently with identical data splits and training budgets; checkpoints are selected by mean validation DSC.
\begin{table}[t]
    \centering
    \caption{
        Default hyperparameters.
        Merged entries apply to all datasets.
        Per-level vectors are ordered from coarse to fine.
    }
    \label{Tab: S1}

    \footnotesize
    \setlength{\tabcolsep}{3pt}
    \renewcommand{\arraystretch}{1.12}

    \begin{tabular*}{\columnwidth}
        {@{\extracolsep{\fill}}lcc@{}}
        \toprule
        \textbf{Parameter}
        & \textbf{IXI/OASIS}
        & \textbf{AbdomMRCT}
        \\
        \midrule

        $(r_L,\ldots,r_1)$
        & \multicolumn{2}{c}{$(3,2,1)$}
        \\

        $(\tau_L^{(0)},\ldots,\tau_1^{(0)})$
        & \multicolumn{2}{c}{$(0.07,0.07,0.07)$}
        \\

        $\lambda_{\mathrm{ms}}$
        & \multicolumn{2}{c}{0.25}
        \\

        $\lambda_{\mathrm{str}}$
        & 0.05
        & 0
        \\

        $\lambda_{\mathrm{vel}}$
        & \multicolumn{2}{c}{0.02}
        \\

        $\lambda_{\mathrm{inv}}$
        & \multicolumn{2}{c}{0.05}
        \\

        $\lambda_{\mathrm{jac}}$
        & \multicolumn{2}{c}{0.01}
        \\

        $\lambda_{\mathrm{conf}}$
        & \multicolumn{2}{c}{0.05}
        \\

        $t_c$
        & \multicolumn{2}{c}{0.15}
        \\

        \bottomrule
    \end{tabular*}
\end{table}

\subsection{Effect of Correlation Temperature}
\label{sec:supp_temperature}

On IXI, we compare learnable per-level temperatures with fixed
temperatures $\tau_l=s\tau_l^{(0)}$, where
$s\in\{0.5,1,2\}$ and $\tau_l^{(0)}$ denotes the default
initialization reported in
Sec.~\ref{Sec: supp_reproducibility}. Fixed temperatures remain unchanged during training
and inference.

Table~\ref{tab:supp_temperature} reports registration
performance and the mean normalized entropy at the finest
level, $\bar{H}_1$, evaluated over the same image pairs,
directions, and spatial regions.
Per-level entropy statistics are additionally recorded
to examine how temperature learning changes the
correspondence distributions across scales.

\begin{table}[t]
    \centering
    \caption{
        Temperature ablation on IXI.
        $\bar{H}_1$ summarizes distribution concentration
        and has no preferred direction.
    }
    \label{tab:supp_temperature}
    \small
    \setlength{\tabcolsep}{2pt}
    \renewcommand{\arraystretch}{1.1}

    \begin{tabular*}{\columnwidth}
        {@{\extracolsep{\fill}}lcccc@{}}
        \toprule
        Temperature
        & DSC $\uparrow$
        & \shortstack{HD95 $\downarrow$\\(mm)}
        & \shortstack{Fold $\downarrow$\\(\%)}
        & $\bar{H}_1$
        \\
        \midrule
        Fixed $0.5\tau_l^{(0)}$
        & 0.7958 & 2.241 & 0.0046 & 0.412 \\
        Fixed $\tau_l^{(0)}$
        & 0.8026 & 2.134 & 0.0029 & 0.684 \\
        Fixed $2\tau_l^{(0)}$
        & 0.7931 & 2.285 & 0.0014 & 0.876 \\
        Learned (default)
        & 0.8093 & 2.040 & 0.0032 & 0.618 \\
        \bottomrule
    \end{tabular*}
\end{table}

\section{Extended Comparisons}
\subsection{Additional Baseline Comparisons}
\label{sec:supp_additional_baselines}
\begin{figure}[t]
    \centering
    \includegraphics[width=0.47\textwidth]{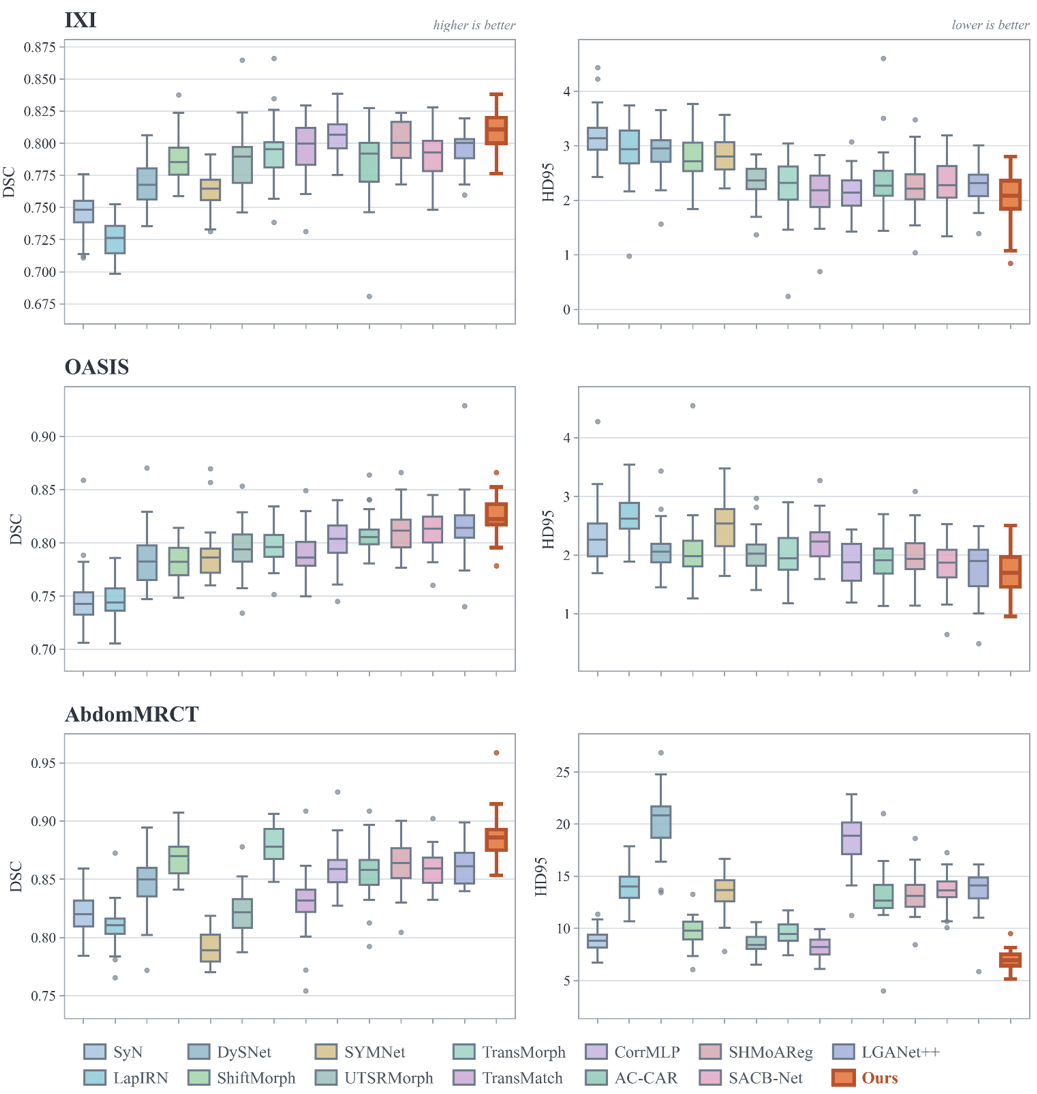}
    \caption{
      Comparison of DSC and HD95 distributions across registration methods.
    }
    \label{Fig: sup_box}
\end{figure}

\begin{figure*}[t]
    \centering
    \includegraphics[width=0.9\textwidth]{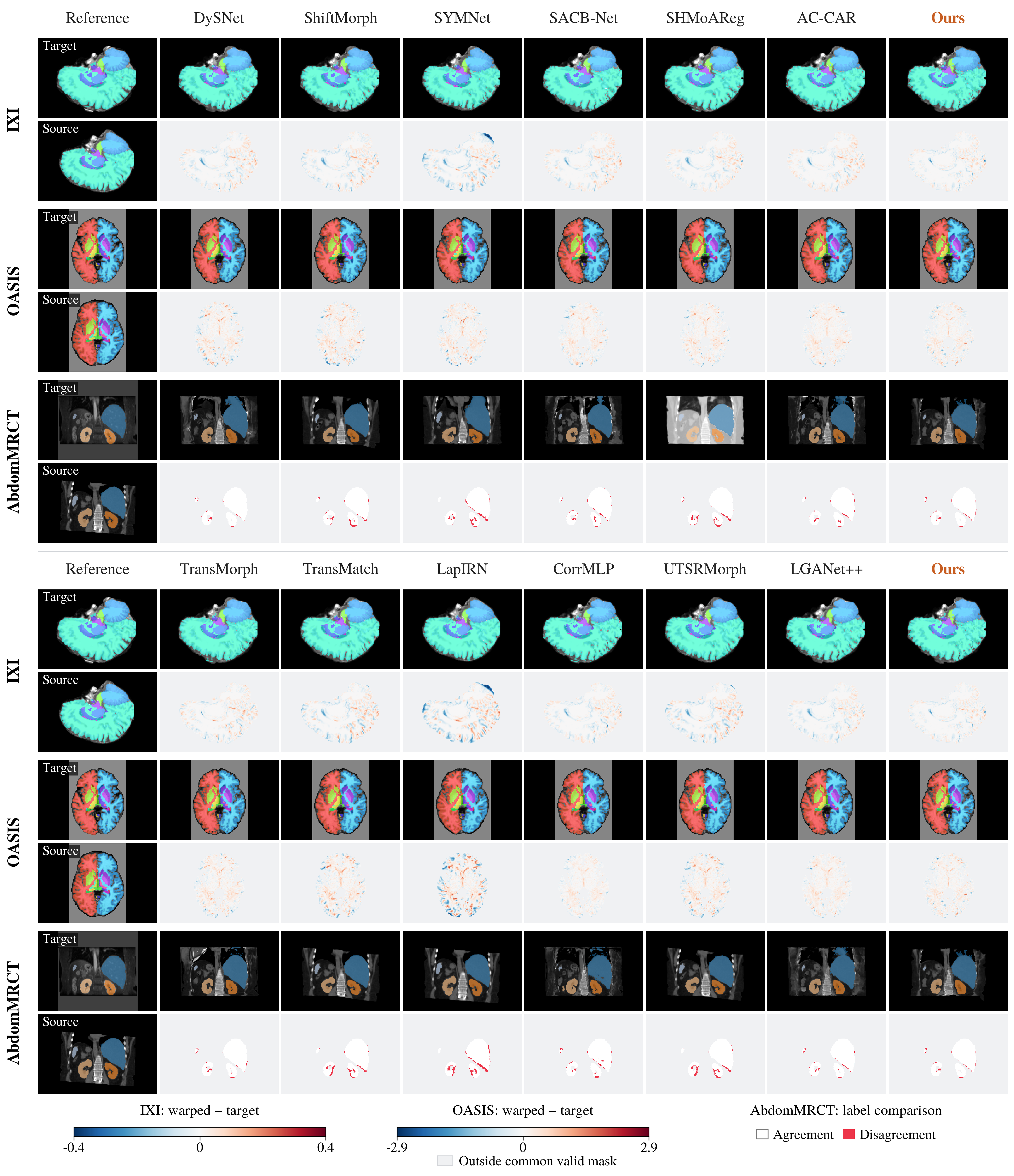}
    \caption{
      Qualitative registration comparison on IXI, OASIS, and AbdomMRCT. For each dataset, method columns show warped images with segmentation overlays (top) and signed intensity residuals for IXI/OASIS or label disagreement maps for AbdomMRCT (bottom).
    }
    \label{Fig: extend_results}
\end{figure*}

\begin{figure}[t]
    \centering
    \includegraphics[width=0.47\textwidth]{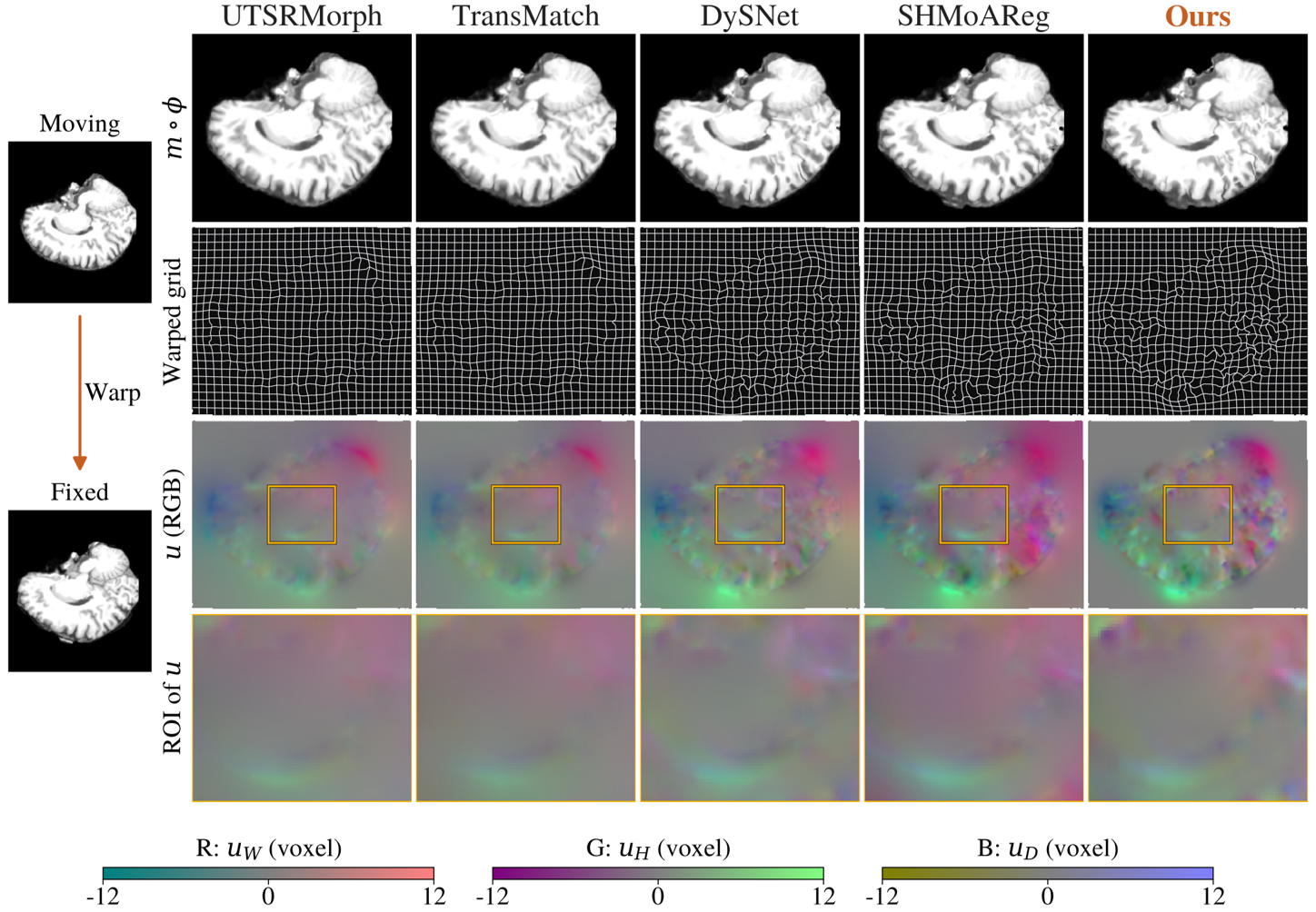}
    \caption{
      Qualitative comparison of IXI registration. 
    }
    \label{Fig: sup_IXI}
\end{figure}

Table~\ref{Tab: supp_comparison} reports the quantitative results for the additional baselines, including Jaccard scores alongside DSC, HD95, and deformation folding rates, while Figure~\ref{Fig: sup_box} illustrates the corresponding DSC and HD95 distributions across all compared methods. Furthermore, Figure~\ref{Fig: extend_results} provides extended qualitative comparisons across IXI, OASIS, and AbdomMRCT, and Figure~\ref{Fig: sup_IXI} visualizes the warped grids and displacement fields for the additional baselines on IXI.

\paragraph{Comparison protocol.}
All methods use the same dataset splits, evaluation pairs,
preprocessing, input resolutions, and affine initialization.
The learning-based methods use the designated training
data for each dataset, with segmentation labels excluded
from their training losses.

We retain each baseline's architecture, transformation parameterization, and method-specific geometric regularization terms.

For abdominal CT--MR registration, we replace modality-dependent intensity similarity losses with MIND-SSC descriptor dissimilarity where cross-modal adaptation is required. Methods with native modality-independent matching objectives retain those objectives.

\paragraph{Metrics and aggregation.}
We follow the evaluation and aggregation protocol used in the main paper and additionally report Jaccard scores. Moving segmentation labels are warped to the fixed-image grid using nearest-neighbor interpolation. Jacobian determinants are computed using the same evaluation domain, finite-difference scheme, and boundary handling across methods.

\begin{table*}[t]
\centering
\caption{
Complete registration comparisons on IXI, OASIS,
and AbdomMRCT.
DSC and Jaccard are reported on a $0$--$1$ scale,
HD95 in millimeters, and Fold as a percentage.
Values are averaged over the evaluated image pairs.
All results are obtained on the held-out test pairs.
}
\label{Tab: supp_comparison}

\footnotesize
\setlength{\tabcolsep}{3pt}
\renewcommand{\arraystretch}{1.10}

\resizebox{\textwidth}{!}{
\begin{tabular}{@{}l*{12}{c}@{}}
\toprule

& \multicolumn{4}{c}{IXI}
& \multicolumn{4}{c}{OASIS}
& \multicolumn{4}{c}{AbdomMRCT} \\

\cmidrule(lr){2-5}
\cmidrule(lr){6-9}
\cmidrule(lr){10-13}

Method
& DSC $\uparrow$
& Jaccard $\uparrow$
& HD95 $\downarrow$
& Fold $\downarrow$
& DSC $\uparrow$
& Jaccard $\uparrow$
& HD95 $\downarrow$
& Fold $\downarrow$
& DSC $\uparrow$
& Jaccard $\uparrow$
& HD95 $\downarrow$
& Fold $\downarrow$ \\

\midrule

UTSRMorph
& 0.7867 & 0.6552 & 2.346 & 0.0320
& 0.7943 & 0.6822 & 2.012 & 0.0709
& 0.8215 & 0.7038 & 8.528 & 0 \\

TransMatch
& 0.7966 & 0.6673 & \underline{2.155} & 0.0136
& 0.7886 & 0.6756 & 2.203 & 0.0527
& 0.8303 & 0.7162 & \underline{8.136} & 0 \\

DySNet
& 0.7683 & 0.6308 & 2.889 & 0.0582
& 0.7845 & 0.6654 & 2.084 & 0.0739
& 0.8476 & 0.7404 & 20.402 & 8.2127 \\

SHMoAReg
& \underline{0.8005} & \underline{0.6729} & 2.230 & 0.5839
& 0.8113 & 07017 & 1.950 & 0.7721
& \underline{0.8631} & \underline{0.7614} & 13.33 & 7.7402 \\

\midrule

\textbf{ReliMorph (ours)}
& \textbf{0.8093} & \textbf{0.6845} & \textbf{2.040} & 0.0032
& \textbf{0.8258} & \textbf{0.7222} & \textbf{1.709} & 0.0423
& \textbf{0.8868} & \textbf{0.7894} & \textbf{6.970} & 0.0290 \\

\bottomrule
\end{tabular}
}
\end{table*}

\subsection{Modality-Independent Structural Evaluation}
\label{sec:supp_multimodal_similarity}
To assess cross-modal structural agreement beyond the four annotated organs on AbdomMRCT, we report $\mathcal{L}_{\text{MIND-SSC}}$ in Table~\ref{Tab: reg} of the main paper using the same checkpoints and trilinearly warped CT images $\widetilde{I}_{m,a} = \mathcal{W}(I_m, u_a)$ from the registration comparisons. Specifically, we extract $K=12$ channel MIND-SSC descriptors\cite{MIND-SSC} $D(\cdot, x) \in \mathbb{R}^K$ with a patch radius of two voxels and a dilation of two voxels under identical preprocessing and normalization across methods, and compute \begin{equation}
    \mathcal{L}_{\text{MIND-SSC}} = \frac{1}{K\vert{}\Omega_{\text{eval}}\vert{}} \sum_{x \in \Omega_{\text{eval}}} \Vert{} D(I_f, x) - D(\widetilde{I}_{m,a}, x) \Vert{}_2^2.
\end{equation}

Here, the common evaluation region $\Omega_{\text{eval}}$ is formed by intersecting the fixed-MR body foreground mask (obtained via Otsu thresholding before intensity normalization, largest-connected-component selection, and hole filling) with the valid in-bounds descriptor neighborhoods across all evaluated methods.

\section{Gate and Geometric Regularity}
\noindent\textbf{Experimental protocol.}
Using the velocity gate defined in Eq.~(\ref{Eq: velocity_modulation}) of the main
paper, we evaluate $c_{\min}\in\{0,0.1,0.25,0.5,1\}$
on IXI and OASIS. The settings $c_{\min}=0$ and
$c_{\min}=1$ remove the positive floor and yield an
identity gate, respectively. The confidence head and its auxiliary
supervision are retained in all configurations, including
$c_{\min}=1$, and the integration depth is fixed at $T=7$.

\noindent\textbf{Evaluation.}
Table~\ref{Tab: gate_floor} summarizes gate-floor sensitivity on IXI and OASIS using DSC, HD95, Fold, and bidirectional
inverse-consistency error (ICE). ICE is the mean Euclidean distance, in millimeters, between a point and its round-trip mapping, averaged over both directions at valid in-bounds sampling locations.

\begin{table}[t]
    \centering
    \caption{
        Gate-floor sensitivity on IXI and OASIS.
        Fold denotes the percentage of voxels with
        $\det(J_{\phi})\leq 0$; ICE denotes bidirectional
        inverse-consistency error.
        The default gate floor is highlighted in bold.
    }
    \label{Tab: gate_floor}

    \small
    \setlength{\tabcolsep}{3pt}
    \renewcommand{\arraystretch}{1.1}

    \begin{tabular*}{\columnwidth}
        {@{\extracolsep{\fill}}lcccc@{}}
        \toprule
        $c_{\min}$
        & DSC $\uparrow$
        & \shortstack{HD95 $\downarrow$\\(mm)}
        & \shortstack{Fold $\downarrow$\\(\%)}
        & \shortstack{ICE $\downarrow$\\(mm)}
        \\
        \midrule

        \multicolumn{5}{c}{\textbf{IXI}} \\
        \midrule
        $0$             & 0.7745 & 2.460 & 0  & 0.6387 \\
        $0.1$           & 0.7732 & 2.497 & 0  & 0.4745 \\
        $\mathbf{0.25}$ & 0.8093 & 2.040 & 0.0032 & 0.4412     \\
        $0.5$           & 0.7875 & 2.456 & 0  & 0.5126 \\
        $1$             & 0.7919 & 2.314 & 0.0001 & 0.4748     \\

        \midrule
        \multicolumn{5}{c}{\textbf{OASIS}} \\
        \midrule
        $0$             & 0.7649 & 2.220 & 0.0001 & 0.5993 \\
        $0.1$           & 0.7941 & 1.965 & 0.0007 & 0.5260 \\
        $\mathbf{0.25}$ & 0.8258 & 1.709 & 0.0423 & 0.4618 \\
        $0.5$           & 0.8084 & 1.852 & 0.0186 & 0.5042 \\
        $1$             & 0.8115 & 1.829 & 0.0245 & 0.4895 \\
        \bottomrule
    \end{tabular*}
\end{table}

\noindent\textbf{Results.}
Among the tested settings, $c_{\min}=0.25$ achieves the
highest DSC and lowest HD95 and ICE on both datasets,
but also the highest measured folding rate. This indicates
a trade-off between anatomical alignment and folding.
Because each configuration is retrained, the results
reflect jointly learned velocity and confidence predictions.

\section{Expert and Routing Analysis}
We examine how expert composition and routing policies
affect registration performance, and whether the learned
experts exhibit region-dependent specialization.

\subsection{Effect of Expert Composition}
\label{sec:supp_expert_composition}

\noindent\textbf{Experimental design.}
We investigate whether registration benefits from the individual
expert branches and their heterogeneous spatial sampling patterns.
The default configuration contains a local expert ($E_{\mathrm{loc}}$), an axial expert ($E_{\mathrm{axial}}$), and a dilated expert ($E_{\mathrm{dil}}$).
The expert architectures are fixed, and their parameters are learned jointly with the registration network.

We remove each expert in turn and renormalize the routing weights over the remaining branches. In addition, we evaluate two alternative sets of experts. The homogeneous configuration uses three independently parameterized local experts with identical architectures. The dilation-only configuration uses three experts with $3\times3\times3$ depthwise convolutions and dilation rates of $1$, $2$, and $3$, respectively, followed by the same pointwise projection. These alternatives examine whether multiple branches alone are sufficient and whether the default combination offers benefits over another set of spatial contexts.

\noindent\textbf{Controlled settings.}
On IXI, all configurations use soft routing and identical expert
output dimensionality. The encoder, matching representation,
velocity decoder, confidence gating, and training objective
remain unchanged. The gate floor and integration depth are
fixed at $c_{\min}=0.25$ and $T=7$, respectively.

Table~\ref{tab:supp_expert_composition} compares registration performance, parameter counts, and FLOPs across expert configurations. The former assess branch contributions, whereas the latter probe the role of heterogeneous receptive-field patterns.

\begin{table*}[t]
    \centering
    \caption{
        Effect of expert composition on IXI.
        All configurations use soft routing and are trained
        independently.
        $L$, $A$, and $D$ denote the local, axial, and dilated
        experts, respectively.
        $L_1$, $L_2$, and $L_3$ have identical architectures
        but independent parameters.
        $D_r$ denotes a dilated expert with dilation rate $r$.
    }
    \label{tab:supp_expert_composition}

    \small
    \setlength{\tabcolsep}{7pt}
    \renewcommand{\arraystretch}{1.15}

    \begin{tabular}{@{}llccccc@{}}
        \toprule
        \textbf{Configuration}
        & \textbf{Expert set}
        & DSC $\uparrow$
        & HD95 $\downarrow$
        & Fold $\downarrow$
        & Params
        & FLOPs
        \\

        &
        &
        & (mm)
        & (\%)
        & (M)
        & (G)
        \\
        \midrule

        Full model
        & $\{L,A,D\}$
        & 0.8093 & 2.040 & 0.0032 & 0.431 & 123.61
        \\

        w/o local expert
        & $\{A,D\}$
        & 0.7890 & 2.343 & 0 & 0.427 & 123.612
        \\

        w/o axial expert
        & $\{L,D\}$
        & 0.7878 & 2.341 & 0 & 0.428 & 123.713
        \\

        w/o dilated expert
        & $\{L,A\}$
        & 0.7891 & 2.330 & 0.0007 & 0.422 & 123.612
        \\

        Homogeneous experts
        & $\{L_1,L_2,L_3\}$
        & 0.7873 & 2.361 & 0.0001 & 0.431 & 124.335
        \\

        Dilation-only experts
        & $\{D_1,D_2,D_3\}$
        & 0.7809 & 2.413 & 0 & 0.431 & 124.335
        \\

        \bottomrule
    \end{tabular}
\end{table*}


\subsection{Comparison of Routing Policies}
\label{sec:supp_routing_policies}

\noindent\textbf{Routing alternatives.}
We compare uniform fusion, soft routing, top-1 routing,
and top-2 routing using the default three-expert configuration.
Uniform fusion assigns a weight of $1/3$ to each expert.
Soft routing uses the complete softmax distribution,
allowing all three experts to contribute at each voxel.
Top-$k$ routing retains only the $k$ largest routing weights
and renormalizes them over the selected experts.

Let $\alpha_{l,e}(p)$ denote the soft routing weight of
expert $e$ at location $p$ and pyramid level $l$.
For $k\in\{1,2\}$, let $\mathcal{K}_{l,k}(p)$ be the set
of indices corresponding to the $k$ largest weights.
The selected weights are
\begin{equation}
    \widetilde{\alpha}_{l,e}^{(k)}(p)
    =
    \frac{
        \alpha_{l,e}(p)
        \mathbf{1}\!\left[e\in\mathcal{K}_{l,k}(p)\right]
    }{
        \sum_j
        \alpha_{l,j}(p)
        \mathbf{1}\!\left[j\in\mathcal{K}_{l,k}(p)\right]
    },
    \label{eq:supp_topk_routing}
\end{equation}
where $\mathbf{1}[\cdot]$ is the indicator function.

\noindent\textbf{Training and inference.}
For top-1 and top-2 routing, we use a straight-through
estimator:
\begin{equation}
    \alpha_{l}^{\mathrm{ST},(k)}
    =
    \alpha_l
    +
    \operatorname{sg}
    \left(
        \widetilde{\alpha}_{l}^{(k)}-\alpha_l
    \right),
    \label{eq:supp_routing_ste}
\end{equation}
where $\operatorname{sg}$ stops gradients. The forward pass uses the selected weights, while gradients with respect to the routing logits are propagated through the soft weights. At inference, we directly use the selected and renormalized weights defined in
Eq.~\eqref{eq:supp_topk_routing}.

All routing policies use the same expert architectures,
confidence-gate design, and training objective. No additional
routing-specific auxiliary losses are used during training.

\noindent\textbf{Accuracy and computational cost.}
Table~\ref{tab:supp_routing_policies} reports registration accuracy, folding, inference time, and peak allocated GPU memory. Timing and memory measurements follow the protocol in Sec.~\ref{Sec: A6}. All expert feature maps are computed before routing weights are applied, so top-$k$ selection changes feature fusion without skipping expert convolutions.

\subsection{Regional Expert Specialization}
\label{Sec:supp_expert_specialization}

\noindent\textbf{Ambiguity stratification.}For each evaluation pair, we trilinearly resample the finest-level pre-update normalized entropy \(H_1\) and routing-weight maps to the fixed-image grid. Within the fixed-image anatomical foreground, the one-third and two-thirds entropy quantiles define the low-, medium-, and high-ambiguity groups. Group membership is determined from the unmodified full-model pass and retained across all expert interventions. Regional statistics are computed separately for each image pair and then averaged across pairs.

\noindent\textbf{Frozen-model expert suppression.}We suppress each expert separately at the finest matching level while keeping all trained parameters and coarser-level computations fixed. The selected expert receives zero weight, and the remaining weights are renormalized to sum to one. We then recompute the final transformation and measure the change in anatomical label disagreement within each fixed ambiguity group. In contrast, the expert-composition variants in Table S5 are trained independently.

\begin{table}[t]
    \centering
    \caption{Comparison of routing policies on IXI using the same local, axial, and dilated experts. Inference time and peak GPU memory are measured for full bidirectional inference at batch size one.
    }
    \label{tab:supp_routing_policies}

    \footnotesize
    \setlength{\tabcolsep}{2pt}
    \renewcommand{\arraystretch}{1.12}

    \begin{tabular*}{\columnwidth}
        {@{\extracolsep{\fill}}lccccc@{}}
        \toprule
        \textbf{Policy}
        & DSC $\uparrow$
        & HD95 $\downarrow$
        & Fold $\downarrow$
        & Time $\downarrow$
        & Mem. $\downarrow$
        \\
        & & (mm) & (\%) & (s/pair) & (GB)
        \\
        \midrule

        Uniform
        & 0.7907
        & 2.312
        & 0.0001
        & $0.77 \pm 0.12$
        & 1.73
        \\

        Soft
        & 0.8093
        & 2.040
        & 0.0032
        & $0.78 \pm 0.12$
        & 1.75
        \\

        Top-1
        & 0.7942
        & 2.264
        & 0.0048
        & $0.79 \pm 0.12$
        & 1.76
        \\

        Top-2
        & 0.8018
        & 2.151
        & 0.0039
        & $0.79 \pm 0.12$
        & 1.76
        \\

        \bottomrule
    \end{tabular*}
\end{table}

\section{Reliability Evaluation}
\label{sec:supp_reliability}
We evaluate how confidence supervision and velocity modulation affect registration, how confidence scores relate to anatomical label disagreement, and whether they identify low-quality registration pairs.

All experiments use the same IXI data splits and evaluation pairs as the main paper. We evaluate confidence scores consistently in the moving-to-fixed registration direction. For pair-level quality assessment, thresholds are determined on validation data and applied unchanged to the held-out test set.

\subsection{Confidence Supervision and Velocity Modulation}
\label{sec:supp_conf_supervision}

\noindent\textbf{Controlled training comparisons.}
We first compare the full model with a variant trained without
confidence supervision by setting $\lambda_{\mathrm{conf}}=0$.
The confidence head and velocity gate are retained, so the
modulation scores can still receive gradients through the
registration objective.
This comparison isolates the contribution of the auxiliary
confidence target from the existence of the gating mechanism.

We additionally evaluate a variant with identity velocity
gates, $g_l(p)\equiv1$, at every pyramid level.
Its confidence head and auxiliary supervision are retained,
allowing the effect of velocity modulation to be examined
separately.

\noindent\textbf{Mean-matched constant-gate intervention.}
To distinguish spatially adaptive modulation from uniform
attenuation, we perform an additional intervention on the
trained full model.
For each image pair, pyramid level, and direction, we first
record the mean gate value from an unmodified inference pass:
\begin{equation}
    \overline{g}_{l,d}
    =
    \frac{1}{|\Omega_l|}
    \sum_{p\in\Omega_l}
    g_{l,d}(p),
    \label{eq:supp_mean_gate}
\end{equation}
where $\Omega_l$ is the spatial grid at level $l$ and $d$
denotes the registration direction.
We then repeat inference with all parameters frozen and
replace each spatial gate by the corresponding recorded
constant $\overline{g}_{l,d}$.

This intervention removes spatial variation in the gates while matching their reference mean at each level and direction. Mean velocity magnitudes are not constrained to match between the two inference passes.

\noindent\textbf{Evaluation.}
Table~\ref{tab:supp_conf_supervision} reports anatomical
accuracy, folding, and the area under the risk--coverage curve
(AURC) of the final confidence map.
The definition of AURC is provided in
Sec.~\ref{Sec: supp_conf_error}.
Registration metrics and score quality are assessed jointly,
because an auxiliary objective may affect confidence quality
and deformation accuracy differently.

\begin{table}[t]
    \centering
    \caption{
        Confidence supervision and velocity modulation on IXI.
        The first three variants are trained independently;
        $^\dagger$ denotes a frozen full-model intervention
        with mean-matched constant gates.
        Fold: voxels with $\det(J_{\phi})\leq0$.
        AURC evaluates final confidence against foreground
        anatomical label disagreement.
    }
    \label{tab:supp_conf_supervision}

    \footnotesize
    \setlength{\tabcolsep}{3pt}
    \renewcommand{\arraystretch}{1.12}

    \begin{tabular*}{\columnwidth}
        {@{\extracolsep{\fill}}lcccc@{}}
        \toprule
        \textbf{Variant}
        & DSC $\uparrow$
        & HD95 $\downarrow$
        & Fold $\downarrow$
        & AURC $\downarrow$
        \\
        &
        &
        & (mm)
        & (\%)
        \\
        \midrule

        Full model
        & 0.8093 & 2.040 & 0.0032 & 0.0642
        \\

        w/o conf. supervision
        & 0.7984 & 2.208 & 0.0024 & 0.1185
        \\

        Identity gates
        & 0.7919 & 2.314 & 0.0001 & 0.0718
        \\

        \midrule

        Constant gates$^\dagger$
        & 0.7956 & 2.254 & 0.0019 & 0.0689
        \\

        \bottomrule
    \end{tabular*}
\end{table}


\subsection{Confidence and Registration Error}
\label{Sec: supp_conf_error}

\noindent\textbf{Independent anatomical error.}
We evaluate confidence within the fixed-image anatomical foreground as the primary region, and additionally analyze anatomical boundary voxels separately. Here, a boundary voxel is defined as a foreground voxel with at least one six-connected neighbor carrying a different anatomical label. Let $z(p) \in \{0, 1\}$ denote the binary indicator of anatomical label disagreement at voxel $p$, computed from the unmodified full-model transformation. This anatomical evaluation signal is distinct from the local image-dissimilarity target used for confidence supervision.

\noindent\textbf{Compared reliability scores.}
We compare the raw finest-level confidence score with the
final consistency-adjusted score and three directly available
registration cues.
Writing $\mathcal{U}$ for resampling to the fixed-image grid,
the evaluated scores are
\begin{align}
    s_{\mathrm{raw}}
    &=
    \mathcal{U}(c_{1,d}),
    \label{eq:supp_score_raw}
    \\
    s_{\mathrm{adj}}
    &=
    \mathcal{U}
    \left[
        c_{1,d}
        \left(1-E_{\mathrm{inv},d}^{1,+}\right)
    \right],
    \label{eq:supp_score_adjusted}
    \\
    s_{\mathrm{inv}}
    &=
    \mathcal{U}
    \left[
        1-E_{\mathrm{inv},d}^{1,+}
    \right],
    \label{eq:supp_score_inverse}
    \\
    s_{\mathrm{ent}}
    &=
    1-\mathcal{U}(H_{1,d}),
    \label{eq:supp_score_entropy}
    \\
    s_{\mathrm{res}}
    &=
    \exp\left(-e_d/t_c\right).
    \label{eq:supp_score_residual}
\end{align}
Here, $H_{1,d}$ is the normalized matching entropy at the
finest matching level before its residual update,
$E_{\mathrm{inv},d}^{1,+}$ is the normalized and clipped
inverse-inconsistency map after that update, and $e_d$ is
the local dissimilarity computed from the final registered
images.
The value of $t_c$ is identical to that used during training.

\noindent\textbf{Error detection.}
For each reliability score $s$, we use $1-s$ as the
error-detection score and treat $z=1$ as the positive class.
We report the area under the receiver operating characteristic
curve (AUROC) and average precision (AP).
Metrics are computed separately for each image pair and then
averaged across the evaluation set.
AP is reported together with the error prevalence because
its interpretation depends on the proportion of erroneous
voxels.

\noindent\textbf{Risk--coverage analysis.}
We additionally assess whether higher-confidence regions
contain fewer anatomical errors.
For pair $i$, let $\pi_i$ order its $N_i$ evaluation voxels
by decreasing confidence.
The risk at coverage $k/N_i$ is
\begin{equation}
    R_i\left(\frac{k}{N_i}\right)
    =
    \frac{1}{k}
    \sum_{j=1}^{k}
    z_i\left(\pi_i(j)\right).
    \label{eq:supp_conf_risk}
\end{equation}
The discrete area under this curve is
\begin{equation}
    \mathrm{AURC}_i
    =
    \frac{1}{N_i}
    \sum_{k=1}^{N_i}
    R_i\left(\frac{k}{N_i}\right).
    \label{eq:supp_conf_aurc}
\end{equation}
Lower AURC indicates that the score places fewer erroneous voxels among its high-confidence predictions. Tied scores are resolved using a fixed random ordering sharedacross the compared score definitions. Table~\ref{tab:supp_conf_error} reports error-detection and risk--coverage metrics for the foreground and boundary regions.

\begin{table*}[t]
    \centering
    \caption{Voxel-level error detection and risk--coverage evaluation of reliability scores on IXI using the same full-model transformations. Voxels with anatomical label disagreement form the positive class for AUROC and AP.
    }
    \label{tab:supp_conf_error}

    \small
    \setlength{\tabcolsep}{7pt}
    \renewcommand{\arraystretch}{1.15}

    \begin{tabular}{@{}lcccccc@{}}
        \toprule

        &
        \multicolumn{3}{c}{\textbf{Foreground}}
        &
        \multicolumn{3}{c}{\textbf{Anatomical boundary}}
        \\
        \cmidrule(lr){2-4}
        \cmidrule(l){5-7}

        \textbf{Reliability score}
        & AUROC $\uparrow$
        & AP $\uparrow$
        & AURC $\downarrow$
        & AUROC $\uparrow$
        & AP $\uparrow$
        & AURC $\downarrow$
        \\
        \midrule

        Raw confidence $s_{\mathrm{raw}}$
        & 0.796 & 0.384 & 0.0695 & 0.742 & 0.578 & 0.2184
        \\

        Adjusted confidence $s_{\mathrm{adj}}$
        & 0.824 & 0.432 & 0.0642 & 0.768 & 0.619 & 0.2065
        \\

        Inverse consistency $s_{\mathrm{inv}}$
        & 0.718 & 0.289 & 0.0894 & 0.605 & 0.492 & 0.2562
        \\

        Matching entropy $s_{\mathrm{ent}}$
        & 0.684 & 0.246 & 0.0982 & 0.691 & 0.518 & 0.2448
        \\

        Local dissimilarity $s_{\mathrm{res}}$
        & 0.765 & 0.341 & 0.0768 & 0.719 & 0.546 & 0.2310
        \\

        \bottomrule
    \end{tabular}
\end{table*}

%
%

\subsection{Inference-Time Quality Assessment}
\label{sec:supp_conf_quality}

\noindent\textbf{Pair-level reliability.}
We investigate whether the confidence output can identify
registration pairs that require further inspection.
The detection score must be computable without anatomical
labels at inference. We aggregate voxelwise reliability scores within a foreground mask derived from the fixed MRI.

For this experiment, the mask is obtained from the fixed MR
image before intensity normalization using Otsu thresholding,
followed by largest-connected-component selection and binary
hole filling.
The same mask is used for all compared reliability scores.
For image pair $i$ and score definition $s$, the pair-level
reliability is
\begin{equation}
    q_{i,s}
    =
    Q_{0.10}
    \left(
        \left\{
            s_i(p):
            p\in\Omega_i^{\mathrm{img}}
        \right\}
    \right),
    \label{eq:supp_pair_reliability}
\end{equation}
where $Q_{0.10}$ denotes the tenth percentile and
$\Omega_i^{\mathrm{img}}$ is the image-derived foreground.
This lower-tail statistic summarizes regions receiving
relatively low reliability scores.
The aggregation rule is fixed before evaluation.

\noindent\textbf{Definition of low-quality registration.}
For evaluation purposes, a pair is designated low quality
when its mean anatomical DSC falls below a threshold
$\delta_{\mathrm{DSC}}$:
\begin{equation}
    y_i
    =
    \mathbf{1}
    \left[
        \mathrm{DSC}_i < \delta_{\mathrm{DSC}}
    \right].
    \label{eq:supp_low_quality_event}
\end{equation}
We set $\delta_{\mathrm{DSC}}$ to the lower quartile of
the full model's validation DSC distribution and then
freeze it for evaluation.
This defines low quality relative to the validation cohort.
The same low-quality labels are used when comparing all
score definitions.

Anatomical labels define the reference low-quality registration pairs used to evaluate detection performance.

\noindent\textbf{Validation-based flagging.}
For each score definition, we determine a flagging threshold
from its validation pair-level scores:
\begin{equation}
    \theta_s
    =
    Q_{\eta}
    \left(
        \left\{
            q_{i,s}^{\mathrm{val}}
        \right\}
    \right),
    \qquad
    \eta=0.10.
    \label{eq:supp_flag_threshold}
\end{equation}
An evaluation pair is flagged when
$q_{i,s}<\theta_s$.
The threshold targets approximately ten percent of validation
pairs and is applied unchanged to the evaluation set.

\noindent\textbf{Evaluation metrics.}
We report pair-level AUROC and AP using $-q_{i,s}$ as the low-quality detection score. At the validation-selected operating point, we additionally report the flag rate, low-quality recall, and flag precision.

Table~\ref{Tab: conf_quality} compares the final confidence
map with the raw confidence and directly available reliability
cues.
Any accuracy statistics on retained pairs should be accompanied
by their coverage, so that quality-based selection is
distinguished from improvements to the registration model
itself.

\begin{table}[t]
    \centering
    \caption{
        Inference-time detection of low-quality pairs on IXI.
        The DSC cutoff is determined on validation data and
        fixed during evaluation.
        Score thresholds target a 10\% validation flag rate;
        evaluation flag rates are reported.
        Recall and precision refer to the low-quality class.
    }
    \label{Tab: conf_quality}

    \footnotesize
    \setlength{\tabcolsep}{2pt}
    \renewcommand{\arraystretch}{1.12}

    \begin{tabular*}{\columnwidth}
        {@{\extracolsep{\fill}}lccccc@{}}
        \toprule
        \textbf{Score}
        & AUROC $\uparrow$
        & AP $\uparrow$
        & Flag
        & Rec. $\uparrow$
        & Prec. $\uparrow$
        \\
        & & & (\%) & (\%) & (\%)
        \\
        \midrule

        Raw conf. $s_{\mathrm{raw}}$
        & 0.858
        & 0.672
        & 10.4
        & 35.7
        & 83.3
        \\

        Adj. conf. $s_{\mathrm{adj}}$
        & 0.894
        & 0.746
        & 10.4
        & 39.3
        & 91.7
        \\

        Inv. cons. $s_{\mathrm{inv}}$
        & 0.762
        & 0.528
        & 9.6
        & 25.0
        & 53.8
        \\

        Entropy $s_{\mathrm{ent}}$
        & 0.715
        & 0.464
        & 11.3
        & 25.0
        & 53.8
        \\

        Local dissim. $s_{\mathrm{res}}$
        & 0.819
        & 0.605
        & 10.4
        & 32.4
        & 75.0
        \\

        \bottomrule
    \end{tabular*}
\end{table}

Table~\ref{Tab: conf_quality} shows that
consistency-adjusted confidence achieves the highest
AUROC and AP among the evaluated scores.
The improvement over raw confidence supports
incorporating inverse-consistency information
into pair-level quality assessment.

\section{Inference Efficiency and Memory Analysis}
\label{Sec: A6}
Benchmarking protocol and results. Table~\ref{Tab: supp_efficiency} compares the end-to-end inference latency and peak GPU memory across all baseline methods on IXI ($160 \times 192 \times 224$). All neural models are benchmarked individually in evaluation mode (batch size 1, FP32 without automatic mixed precision) on a single NVIDIA A6000 GPU (48 GB), while SyN is evaluated on the same workstation using a fixed CPU thread count. Latency covers the complete pipeline from a preprocessed image pair to the final full-resolution transformation and warped moving image; for ReliMorph, this includes feature extraction, correlation volumes, reliability statistics, expert routing, confidence gating, and full bidirectional refinement without dividing by the number of directions. After 20 warm-up passes, each test pair is timed over five passes with CUDA synchronization before and after each run, and we report the mean and standard deviation across pairs. Peak GPU memory is recorded as the maximum allocation from the PyTorch CUDA allocator after resetting memory statistics before each pass.

\begin{table}[t]
\centering
\caption{Inference efficiency on IXI with an input size of $160\times192\times224$ and batch size one. Time is reported as mean $\pm$ standard deviation across image pairs. GPU memory is the largest observed peak allocated memory.}
\label{Tab: supp_efficiency}
\small
\setlength{\tabcolsep}{3pt}
\renewcommand{\arraystretch}{1.08}
\begin{tabular*}{\columnwidth}{@{\extracolsep{\fill}}lccc@{}}
\toprule
Method & Device & \shortstack{Time $\downarrow$\\(s/pair)} & \shortstack{GPU memory $\downarrow$\\(GiB)} \\
\midrule
SyN & CPU & $151.6\pm12.3$ & N/A \\
\midrule
TransMorph & GPU & $0.31\pm0.04$ & 4.89 \\
CorrMLP & GPU & $2.39\pm0.22$ & 6.80 \\
AC-CAR & GPU & $0.65\pm0.08$ & 3.66 \\
SACB-Net & GPU & $3.34\pm0.87$ & 2.42 \\
ShiftMorph & GPU & $\boldsymbol{0.07 \pm 0.02}$ & \textbf{1.41} \\
LapIRN (SVF) & GPU & \underline{$0.29\pm0.04$} & 5.59 \\
SYMNet & GPU & $0.58\pm0.06$ & 2.37 \\
LGANet++ & GPU & $2.18\pm0.11$ & 6.37 \\
\midrule
UTSRMorph & GPU & $0.41\pm0.06$ & 5.74 \\
TransMatch & GPU & $0.49\pm0.06$ & 4.02 \\
DySNet & GPU & $0.88\pm0.10$ & 3.27 \\
SHMoAReg & GPU & $0.269\pm0.01$ & 7.29 \\
\midrule
\textbf{ReliMorph (ours)} & GPU & $0.78\pm0.12$ & \underline{1.75} \\
\bottomrule
\end{tabular*}
\end{table}



\end{document}